\documentclass[journal,twoside,web]{ieeecolor}
\usepackage{generic}
\usepackage{cite}
\usepackage{amsmath,amssymb,amsfonts}
\usepackage{algorithmic}
\usepackage{graphicx}
\usepackage{algorithm,algorithmic}
\usepackage{hyperref}
\usepackage{multirow}
\usepackage{subcaption}
\hypersetup{hidelinks=true}
\usepackage{textcomp}
\usepackage{booktabs}
\newcommand{\chingyi}[1]{{#1}}

\def\BibTeX{{\rm B\kern-.05em{\sc i\kern-.025em b}\kern-.08em
    T\kern-.1667em\lower.7ex\hbox{E}\kern-.125emX}}
\begin{document}
\title{ReRAM-aware Model Finetuning addressing I-V Non-linearity and Retention Errors}
\author{Ching-Yi Lin, \IEEEmembership{Member, IEEE}, 
Shamik Kundu, 
Arnab Raha, 
and Sahil Shah,~\IEEEmembership{Senior Member, IEEE}
\thanks{Manuscript submitted for review on [date].}
\thanks{Ching-Yi Lin and Sahil Shah are with the University of Maryland, College Park, MD, USA (e-mail: chingyil@umd.edu and sshah389@umd.edu ).}
\thanks{Shamik Kundu and Arnab Raha are with Intel Corporation, 2200 Mission College Blvd, Santa Clara, CA 95054, USA (e-mail: shamik.kundu@intel.com,arnab.raha@intel.com).}
}

\maketitle

\begin{abstract}
Traditional CPU, GPU, and NPU architectures are increasingly limited by the von Neumann bottleneck. While In-Memory Computing (IMC) using ReRAM crossbar arrays offers a high-density, energy-efficient alternative, its practical deployment is constrained through their non-idealities. Existing hardware-aware training frameworks often require training from scratch, which is computationally prohibitive for modern large-scale models. In this work, we propose a finetuning-based hardware-aware training algorithm that enables robust DNN deployment on ReRAM with minimal training overhead. Our approach mitigates I-V non-linearity by applying a range-shrunk sinh transformation and incorporates retention errors directly into a regularization loss during the finetuning process. We evaluate our framework across models and tasks such as image classification and question-answering (QA). Experimental results demonstrate that our method achieves similar accuracy on large-scale models like ResNet18 and DeiT-Tiny as the base model. In-case of ImageNet for MobileNetV3 families the technique has only \chingyi{less than} 2\% accuracy degradation. Further, applying the technique on the SQuAD v2 dataset results in only 1 point degradation of \chingyi{F-1} score. 
\end{abstract}


\section{Introduction}
\label{sec:introduction}

The development and widespread deployment of large language models across diverse applications has driven increasing demand for localized, low-power DNN inference. Although traditional CPU and GPU architectures have achieved substantial performance improvements, they remain constrained by the \chingyi{\textit{von Neumann bottleneck}}, where frequent data movement between memory and compute units results in significant latency and energy overheads \cite{horowitz_11_2014}. Neural Processing Units (NPUs) offer a more efficient alternative for accelerating machine learning workloads \cite{raha_llm-npu_2025}; however, further improvements in energy efficiency require minimizing the physical and architectural separation between memory and computation.

In-memory computing (IMC) has emerged as a promising alternative, enabling weights to be stored and processed directly within the cell array and bypasses weight-related data transfers\cite{baek_edge_2025,verma_-memory_2019,sebastian_memory_2020}. Within the in-memory computing, ReRAM is commonly used for its high density and high energy efficiency \cite{ielmini_-memory_2018,yang_efficient_2025}. 

\begin{figure}[!b]
\centerline{\includegraphics[width=\columnwidth]{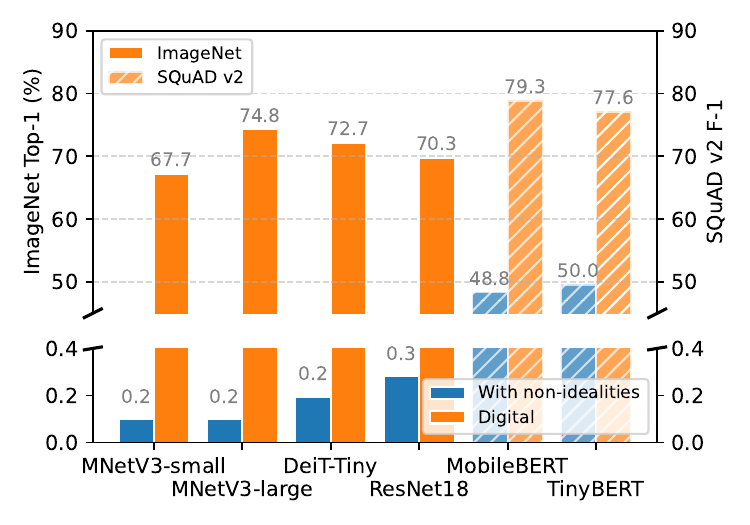}}
\caption{Model performance with and without ReRAM non-idealities\chingyi{, including IV-nonideality and retention error.}}
\label{fig:intro}
\end{figure}

Despite these advantages in ReRAM, the non-idealities in ReRAM in-memory computing have been a large concern in their deployment\cite{didin_characterization_2026}. The primary non-idealities in ReRAM arrays include IR drop, IV non-linearity, retention error, and device-to-device (D2D) mismatch. Among these, IR drop is largely architectural dependent; existing research has mitigated its impact through parasitic-aware layouts and runtime compensation~\cite{huang2021efficient, na2016offset}. Similarly, D2D mismatch can be addressed through common centroid layout techniques or compensation. Conversely, IV non-linearity and retention error are intrinsic physical properties of ReRAM devices, making them inevitable across all array configurations. Thus this work focus\chingyi{es} on the latter two non-idealities due to their omnipresence. Figure \ref{fig:intro} shows the accuracy with and without these two non-idealities, which has significant effect in accuracy across various models and tasks. By targeting these intrinsic errors, our approach aims to train models robust to non-linear I-V and retention-referred weight shifts. This model-level solution not only decompose non-ideality consideration and architectural design but also remain orthogonal to existing architectural solutions designed to compensate for IR drop and D2D mismatch.

\begin{figure*}[!t]
    \centering
    \includegraphics[width=\linewidth]{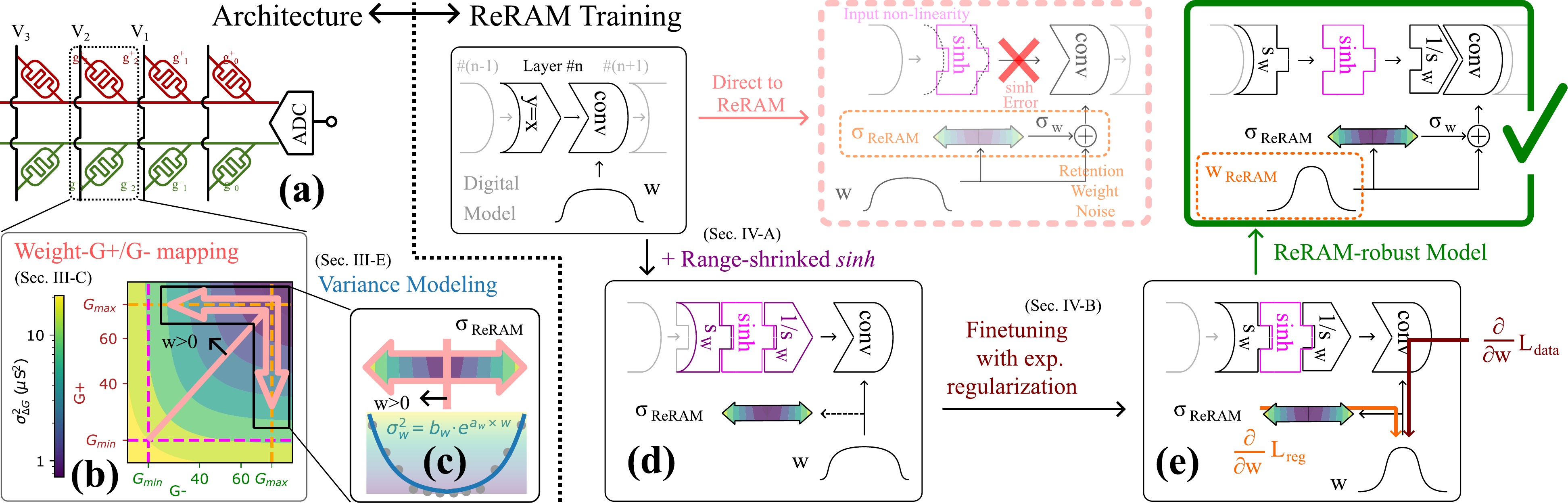}
    \caption{Methods overview in architecture (a) Differential ReRAM architecture (b) Weight-to-$G+/G-$ mapping that minimizes $\sigma_{\Delta G}$ (c) Modeling weight-inferred variance $\sigma_w$ in an exponential relationship and training algorithm to accommodate ReRAM non-ideality (input non-linearity and retention weight error) through (d) Range-shrinked sinh and (e) Finetuning with exponential regularization}
    \label{fig:overview}
\end{figure*}

To consider the hardware non-idealities above, various hardware-aware training frameworks have been proposed, spanning from compute-graph level~\cite{lee2024mitigating,chakraborty2020geniex}, module-level~\cite{rasch2023hardware, joardar2021learning}, to new training algorithm~\cite{wan2022compute}. However, all these works rely on training from scratch, with most evaluations confined to relatively small models. Given the modern shift toward large-scale models and task-specific finetuning, such approaches are often impractical for its expensive training cost or high requirements for training resources. Considering this, our method in this work targets finetuning-based manner to accommodate ReRAM-specific non-idealities. By treating ReRAM deployment as a downstream task, our flow aligns with the current LLM development, enabling the transition of large-scale foundational models to efficient edge hardware with minimal training overhead.

\begin{figure*}[t]
    \centering
    \includegraphics[width=\linewidth]{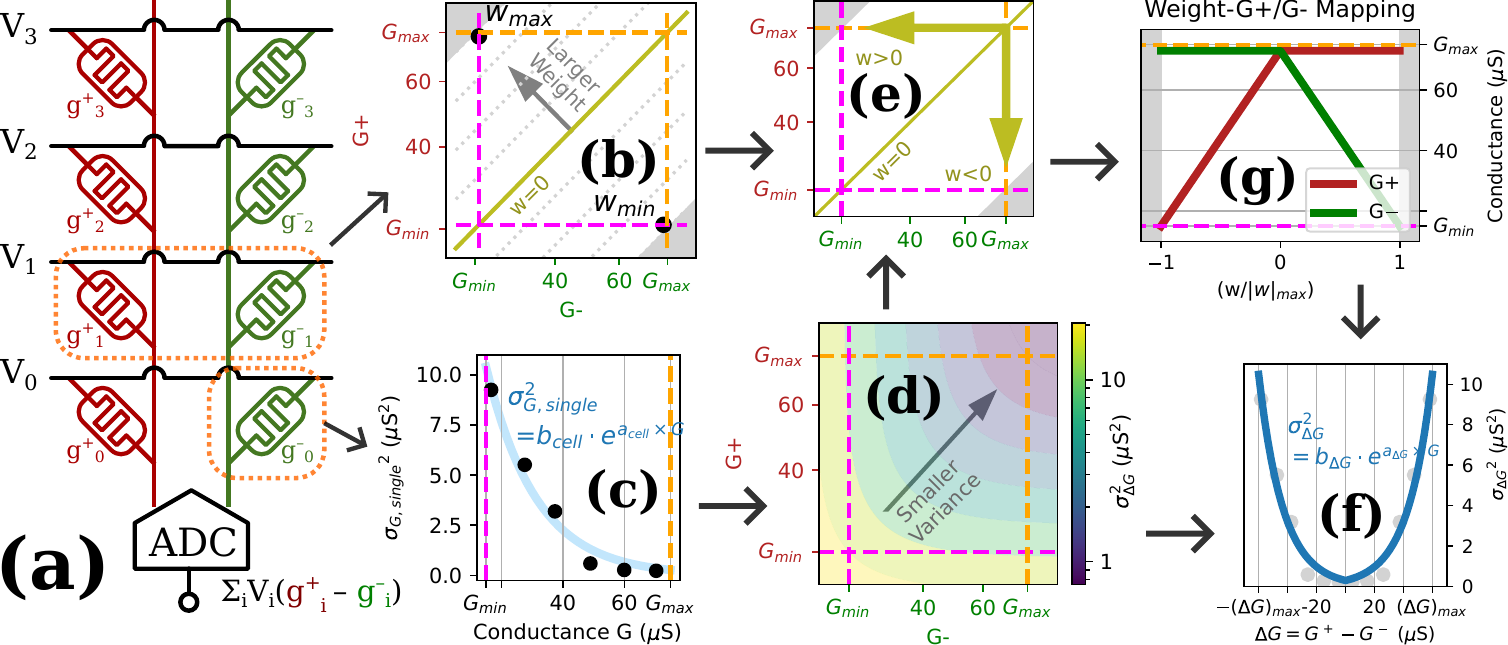}
    \caption{\chingyi{(a) Differential ReRAM architecture (b) Weight representation in the $G^+$-$G^-$ space with lower-/upper-bound conductance in magenta/orange. Iso-weight lines are parallel to the $w=0$ diagonal (c) Single-cell conductance variance $\sigma^2_{G,single}$-vs-$G$ plot exhibiting a decaying exponential relationship (d) MAC-level conductance variance $\sigma^2_{\Delta G}$ in a $G^+$-$G^-$ plot extended from $\sigma^2_{G,single}(G)$ (e) Mapping strategy of ($G^+$,$G^-$) at different target weight $w$ (f) Resulting weight variance $\sigma^2_{\Delta G}$-vs-$G$ plot with low-variance advantage in smaller weights (g) Weight-to-conductance mapping between target weight $w$ and the conductance in $G+$ and $G-$ cells.}}
    \label{fig:reram-architecture}
\end{figure*}

In this work, we exploit a low-variance conductance state in ReRAM devices and introduce a finetuning technique for models to compensate the accuracy degradation from ReRAM non-idealities. Figure~\ref{fig:overview} illustrates the key components of the proposed framework. On the architecture side, we employ a differential ReRAM architecture together with a weight-to-conductance mapping strategy that minimizes MAC-level variance and models it using an exponential relationship. On the training side, input-transformation nonlinearity is incorporated into pretrained models, while retention-induced weight variance is formulated as a regularization loss jointly optimized with the task loss during model adaptation. We evaluate the proposed method on six models across image classification and question-answering (QA) tasks. The results show iso-accuracy for larger models such as ResNet-18 and DeiT-Tiny, and less than 2\% accuracy degradation for edge models in the MobileNetV3 family.


\section{Background}
\subsection{ReRAM Non-idealities}
ReRAM has emerged as a frontrunner for next-generation non-volatile memory and neuromorphic computing due to its high density, low power consumption, and CMOS compatibility. However, programming ReRAM resistance (conductance) introduces several non-idealities spanning the behavioral, device, and architectural level.

At the behavioral level, the non-linearity between output current and input voltage is widely recognized as a primary source of ReRAM non-ideality. Lentz showed the exponential I-V relationship ~\cite{lentz2013current}, while Messaris further modeled this I-V using $sinh$ and implemented the characteristic in Verilog-A~\cite{messaris2018data}. Beyond I-V non-linearity, device-to-device mismatch also significantly affects their electrical characteristics, causing deviations from nominal behavior and resulting in inconsistent effective resistance under identical programming conditions. This issue has been addressed through adaptive amplitude programming algorithms~\cite{alibart2012high}, and further improved via adaptive pulse duration schemes~\cite{didin_characterization_2026} or write-verify techniques~\cite{gonugondla2020swipe}.

At the device level, ReRAM suffers from both stuck-at faults and long-term retention drifts. Stuck-at faults trap individual cells at fixed resistance states. Although such faults can cause significant accuracy degradation (e.g., over 40\% on the CIFAR-10 dataset~\cite{liu2022online}), they can be effectively fixed through remapping techniques~\cite{nguyen2023craft, shin2022fault}. In contrast, retention degradation leads to drift in resistance values over time. This gradual drift, referred to as \textit{retention error}, has been extensively studied to characterize long-term device behavior of ReRAM.
Kopperberg showed the conductance distribution across 2.8M ReRAM devices~\cite{kopperberg2024accurate}. Lin modeled retention error as a mean shift with a constant rate in the $g$-$log(t)$ domain~\cite{lin2019performance}. Maldonado observed larger increased deviation at lower conductance in through qualitative analysis~\cite{maldonado2025comprehensive}, while Didin further provided quantitative characterization by examining the $\sigma_g$-$g$ relationship \cite{didin_characterization_2026}.

At the architecture level, IR drop and ADC quantization errors are two dominant sources of non-idealities~\cite{rasch2023hardware}. IR drop leads to voltage degradation in large crossbar arrays, while ADC quantization introduces discretization errors. These non-idealities can be mitigated through smaller arrays and higher-precision ADCs. Unlike behavioral- and device-level non-idealities, these architectural non-idealities primarily introduce design trade-offs rather than being fundamentally unavoidable.

In this work, we focus on I-V non-linearity and retention errors, and propose a model-level algorithm to mitigate their impact. Our approarch is orthogonal to existing ReRAM programming algorithms and architectural optimizations. Moreover, model-level solutions can be integrated into existing hardware without requiring new tapeouts, thereby reducing deployment cost and facilitating scalability.

\subsection{Hardware-aware training}
The hardware non-idealities has been known to affect the DNN inference accuracy. Mehonic analyzed MNIST accuracy with discrete conductance states and I-V non-linearity~\cite{mehonic2019simulation}. Based on this, multiple existing works proposed hardware-aware training to address various non-idealities in ReRAM. 

One straightforward style is inject the non-idealities during training. Joksas replaces the linear kernel $w\cdot x$ with a non-linear kernel $g(w,x)$ and draws the parameters from multivariate normal distribution during training~\cite{joksas2022nonideality}. Rasch also inject noise which mimics the non-idealities including dynamic-range limitations, weight-programming errors, PCM drift and system noise~\cite{rasch2023hardware}.

In addition to direct non-ideality injection, some works modify the conventional DL training to make it non-ideality robust. Zhang exploits the quantization-aware training to enhances the robustness of ReRAM accelerators~\cite{zhang2021quantized}. Joardar added weight clipping to prevent stuck-in weights destablizing model training through a positive feedback~\cite{joardar2021learning}. Wan progressively trains the model to by freezing $n-1$ layers and running layer $n$ on ReRAM to obtain the non-ideal output to train $n$-th to the last layer~\cite{wan2022compute}.

Several works translate non-linearities into parameter-modulated terms. Chakraborty exploits NNs to estimate the scale $f_R(G,V)$ to modulate the ReRAM output current emulating SPICE results~\cite{chakraborty2020geniex}. Lee finds the effective weight to reflect the I-V non-linearity and IR drop in ReRAM arrays~\cite{lee2024mitigating}. Chang uses look-up table-based methods to estimate ReRAM outputs from convolution results~\cite{chang2025error}


\begin{figure}[b]
    \centering
    \includegraphics[width=\linewidth]{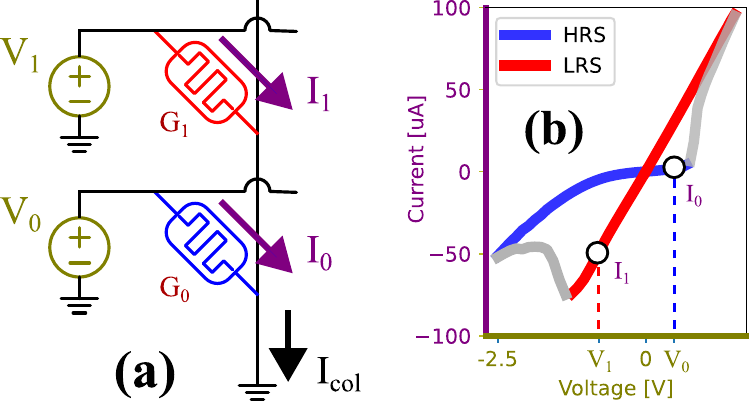}
    \caption{\chingyi{(a) Simple $2\times1$ ReRAM array to compute inner product between voltage input $[V_0,V_1]$ and conductance weight $[G_0,G_1]$(b) ReRAM I-V characteristic~\cite{didin_characterization_2026}}}
    \label{fig:reram-cell}
\end{figure}

\section{ReRAM \chingyi{Devices} and System Architecture}


\subsection{ReRAM cell Behavior}
\label{sec:reram-cell-behavior}
\chingyi{We first illustrate how ReRAM cells perform ML inference. Figure \ref{fig:reram-cell} (a) shows a simple $2\times1$ ReRAM array, where each cell converts input voltage $V_0$ and $V_1$ to output current $I_0$ and $I_1$, respectively. According to Kirchhoff's Current Law (KCL), these partial currents aggregate together to obtain $I_{col}=I_0+I_1$ in a column-based manner.}

\chingyi{Figure \ref{fig:reram-cell}(b) presents the I-V relationship of a ReRAM cell. Through conductive filament and SET/RESET operations, each ReRAM cell can be programmed into different conductance levels. For simplicity, we consider the two extreme states, low-resistance state (LRS) and high-resistance state (HRS), for example.}

\chingyi{In an ideal case, a programmable resistive device should follow Ohm's law, where the output current $I_i$ is the product of programmed conductance $G_i$ and input voltage $V_i$ at the single-cell level. This results in the column current}
\begin{equation*}
I_{col}=\sum_iI_i=\sum_iG_iV_i=\mathbf{G}\cdot\mathbf{V}
\end{equation*}
equals to the inner product between conductance vector $\mathbf{G}$ and input vector $\mathbf{V}$. By mapping model weights $\mathbf{W}$ and inputs $\mathbf{X}$ to $\mathbf{G}$ and $\mathbf{V}$ through appropriate scaling factors $r_{G2w}$ and $r_{v2x}$, the weight-input inner product can be computed by the ReRAM array and a scaling
\begin{equation}
    \mathbf{W}\cdot\mathbf{X} = (r_{G2w}\mathbf{G})\cdot(r_{v2x}\mathbf{V})=(r_{G2w}r_{v2x})(\mathbf{G}\cdot\mathbf{V})
    \label{eqn:wx-gv}
\end{equation}

\chingyi{However, Figure \ref{fig:reram-cell}(b) shows the post-silicon ReRAM measurement results~\cite{didin_characterization_2026}, revealing that the I-V relationship is not ideal linear. Several studies consider the ReRAM output current as a $sinh$ of its conductance-voltage product~\cite{messaris2018data} and make Equation \ref{eqn:wx-gv} no longer holds. To address this challenge, we propose a non-ideality aware training framework that incorporates ReRAM device characteristics into modern fine-tuning-based training pipelines.}
\subsection{Crossbar ReRAM Architecture}\label{SubSec:Diff_ReRAM}
To perform deep learning inference using ReRAM cells, we use a crossbar structure as illustrated in Figure \ref{fig:reram-architecture}(a). In this architecture, each MAC operation is achieved with two ReRAM cells~\cite{chi2016prime,wan2020voltage}, both multiplying the same row-based voltage ($V_i$) with their conductance ($G^+_i$ and $G^-_i$). Its differential partial sum is accumulated in its column and fed into a column-based ADC.

Figure~\ref{fig:reram-architecture}(b) visualizes the mapping from a pair of conductances ($G_i^+, G_i^-$) to their corresponding weights. Any pair with $G_i^+ = G_i^-$ represents a zero weight, as no voltage difference is generated at the ADC; these points lie along the olive-colored line with slope equal to 1. Moving away from this line, parallel lines toward the upper-left region correspond to $G_i^+ - G_i^- > 0$ and represent positive weights, while those toward the lower-right region correspond to $G_i^+ - G_i^- < 0$ and represent negative weights. The conductance values are \chingyi{physically} bounded by two horizontal lines ($G_i^+ = G_{\max}$ and $G_i^+ = G_{\min}$) and two vertical lines ($G_i^- = G_{\max}$ and $G_i^- = G_{\min}$). We highlight $G_{\max}$ and $G_{\min}$ using orange and magenta colors, respectively, for visual clarity across the figure. The maximum weight is achieved at $G_i^+ = G_{\max}$ and $G_i^- = G_{\min}$, located at the \chingyi{top-left} orange--magenta intersection, while the minimum weight occurs at the \chingyi{bottom-right} magenta--orange intersection with the values of $G_i^+$ and $G_i^-$ swapped.

\subsection{Single Time-point Variance Modeling}
\label{sec:var-modeling}
The statistical properties of ReRAM cells have been widely studied, particularly their conductance-dependent variance~\cite{didin_characterization_2026}. Figure~\ref{fig:reram-architecture}(c) shows that ReRAM variation is negatively correlated with conductance. We model the relationship between variance and conductance using an exponential function:

\begin{equation}
    \sigma^2_{G,\text{single}} = b_{\text{cell}} \times \exp(a_{\text{cell}} \times G)
    \label{eqn:exp-model}
\end{equation}

Despite its simplicity, this formulation is quite general. Specifically, Equation~\ref{eqn:exp-model} is equivalent to the form $\sigma^2 = b \times c^{a \times G + d}$, with $b_{\text{cell}} = b \times c^d$ and $a_{\text{cell}} = a \times \ln(c)$.

To extend the single-cell variance model to the differential ReRAM structure described in Section~\ref{SubSec:Diff_ReRAM}, Figure~\ref{fig:reram-architecture}(d) presents a 2-D visualization. The pair-level variance $\sigma^2(G_i^+, G_i^-)$ is estimated as the sum of individual variances, i.e.,
$\sigma^2(G_i^+, G_i^-) = \sigma^2_{G,\text{single}}(G_i^+) + \sigma^2_{G,\text{single}}(G_i^-)$.
This formulation assumes variance additivity, which holds for independent Gaussian random variables. From the color-coded variance in Figure~\ref{fig:reram-architecture}(d), we observe that moving upward (increasing $G_i^+$) or rightward (increasing $G_i^-$) reduces the overall variance. This is due to the negative correlation between conductance and variance at the single-cell level. Consequently, for a fixed conductance difference $\Delta G = G_i^+ - G_i^-$, although multiple $(G_i^+, G_i^-)$ pairs are valid (as shown in Figure~\ref{fig:reram-architecture}(b)), the minimum variance is achieved by maximizing both conductances within the allowed range. In practice, this implies selecting the largest feasible values of $G_i^+$ and $G_i^-$ to minimize the pair-level variance.

Figure~\ref{fig:reram-architecture}(e) illustrates the 1-D realization of $\Delta G$ on the 2-D $G_i^+$--$G_i^-$ plane. This realization lies along the boundary defined by union of $G_i^+ = G_{\max}$ and $G_i^- = G_{\max}$, which minimizes variance for a given $\Delta G$, as motivated by the feasible conductance region in Figure~\ref{fig:reram-architecture}(b) and the variance trends in Figure~\ref{fig:reram-architecture}(d). This leads to two simple constraints: $G_i^+ = G_{\max}$ when $w \geq 0$, and $G_i^- = G_{\max}$ when $w \leq 0$. Although other $(G_i^+, G_i^-)$ combinations within the valid region satisfy the same $\Delta G$, they result in higher variance compared to this boundary realization.

Based on the 1-D $\Delta G$ realization in Figure~\ref{fig:reram-architecture}(e), one device in the differential pair is fixed at $G_{\max}$, while the other is reduced by the magnitude of the desired conductance difference. Thus,
\begin{equation}
    (G_i^+,G_i^-)=
    \begin{cases}
        (G_{\max},\,G_{\max}-\Delta G), & \Delta G \geq 0,\\
        (G_{\max}+\Delta G,\,G_{\max}), & \Delta G < 0.
    \end{cases}
\label{eqn:gpgn-deltag}
\end{equation}
Equivalently, the lower-conductance device is programmed to $G_{\max}-|\Delta G|$. Therefore, assuming independent cell variations, the pairwise conductance-difference variance is
\begin{align}
\begin{split}
    \sigma^2_{\Delta G}
    &= \sigma^2_{G^+} + \sigma^2_{G^-} \\
    &= b_{\text{cell}}\exp(a_{\text{cell}}G_{\max})
    + b_{\text{cell}}\exp\left(a_{\text{cell}}(G_{\max}-|\Delta G|)\right) \\
    &= b_{\text{cell}}\exp(a_{\text{cell}}G_{\max})
    \times\left[1+\exp\left(-a_{\text{cell}}|\Delta G|\right)\right]\\
    &\approx b_{cell} \exp(a_{cell} G_{max})\times exp\left(-a_{\text{cell}}|\Delta G|\right)\\
    &= b_{\Delta G}\times\exp(a_{\Delta G} | \Delta G|).
    \label{eqn:var-g}
\end{split}
\end{align}
where $a_{\Delta G}=-a_{cell}$ and $b_{\Delta G}=b_{\text{cell}}\exp(a_{\text{cell}}G_{\max})$\chingyi{, and the approximation $\exp\left(-a_{\text{cell}}|\Delta G|\right)\gg1$ can be visually justified from the large discrepancy in $\sigma_{G,single}^2$ from Figure \ref{fig:reram-architecture}(c). With $a_{\Delta G}=-a_{cell}>0$}, $\sigma^2_{\Delta G}$ increases with $|\Delta G|$ and is minimized near $\Delta G=0$, producing the convex trend shown in Figure~\ref{fig:reram-architecture}(f) over the valid range $[-(\Delta G)_{\max}, (\Delta G)_{\max}]$, where $(\Delta G)_{\max} = G_{\max} - G_{\min}$.

\subsection{Weight-to-conductance Mapping}

Figure~\ref{fig:reram-architecture}(g) shows the proposed conductance-to-weight mapping for ReRAM arrays. To encode $y_i = x_i \times w_i$ through $\Delta I_i = V \times \Delta G_i$, where $\Delta G_i = G_i^+ - G_i^-$, the weight-to-conductance mapping is \chingyi{derived from Equation \ref{eqn:gpgn-deltag} and} defined as

\begin{equation}
G_i^+ =
\begin{cases}
    G_{\max}, & \text{if } w_i \geq 0 \\
    G_{\max} + \dfrac{w_i}{r_{G2w}}, & \text{otherwise}
\end{cases}
\label{eqn:conductance-mapping1}
\end{equation}
\begin{equation}
G_i^- =
\begin{cases}
    G_{\max}, & \text{if } w_i \leq 0 \\
    G_{\max} - \dfrac{w_i}{r_{G2w}}, & \text{otherwise}
\end{cases}
\label{eqn:conductance-mapping2}
\end{equation}

, where $r_{G2w} = \|w_{\max}\| / (G_{\max} - G_{\min})$ is the conductance-to-weight scaling factor. Equation~\ref{eqn:conductance-mapping1} and \ref{eqn:conductance-mapping2} corresponds to the mapping shown in Figure~\ref{fig:reram-architecture}(g). To build intuition, consider two key points: $w = w_{\max}$ and $w = 0$. For the maximum weight $w = w_{\max}$, the conductances are set to $G_i^+ = G_{\max}$ and $G_i^- = G_{\min}$ to maximize $\Delta G$. In contrast, a zero weight ($w = 0$) is achieved when $G_i^+ = G_i^- = G_{\max}$, resulting in $\Delta G = 0$. This establishes $r_{G2w}$ as the ratio between the conductance range $(G_{\max} - G_{\min})$ and the weight range $\|w_{\max}\|$.

Using this mapping, the variance in the weight domain $\sigma_w^2$ can be derived from the relation $\Delta G = w / r_{G2w}$ and equation for $\sigma^2_{\Delta G}$ shown in \chingyi{Equation} \ref{eqn:var-g}:
\begin{align}
\begin{split}
    \sigma^2_w &= r_{G2w}^2 \cdot \sigma^2_{\Delta G} \\
    &= r_{G2w}^2 \cdot b_{\Delta G} \cdot \exp(a_{\Delta G} \cdot \Delta G) \\
    &= \left[b_{\Delta G} \cdot r_{G2w}^2\right] \cdot \exp\left(\frac{a_{\Delta G}}{r_{G2w}} \cdot w\right) \\
    &= b_w \cdot \exp(a_w \cdot w),
    \label{eqn:sigma-w}
\end{split}
\end{align}
where $a_w = a_{\Delta G} / r_{G2w}$ and $b_w = b_{\Delta G} \cdot r_{G2w}^2$. Notably, $\sigma^2_{G,\text{single}}$, $\sigma^2_{\Delta G}$, and $\sigma^2_w$ all share the same exponential form $b \cdot \exp(a x)$, with different effective parameters (summarized in Table~\ref{tab:coeff}).

In contrast to $\sigma^2_{G,\text{single}}$ and $\sigma^2_{\Delta G}$ that are derived only from device-dependent parameters, $\sigma^2_w$ is affected from $a_{\Delta G}$, $b_{\Delta G}$, and model-level parameters ($w$ and $\|w_{max}\|$). For example,
The prefactor $b_w = b_{\Delta G} \cdot r_{G2w}^2$ represents the baseline variance (i.e., the variance at $w=0$), which is independent of $w$ and stands for the smallest variance according to Figure \ref{fig:reram-architecture} (f). Substituting $r_{G2w} = \|w_{\max}\| / (G_{\max} - G_{\min})$, we obtain
\[
b_w = b_{\Delta G} \cdot \frac{\|w_{\max}\|^2}{(G_{\max} - G_{\min})^2}.
\]
This expression shows that the variance scales quadratically with the weight range and inversely with the conductance range. For a fixed device (i.e., fixed $G_{\max}$ and $G_{\min}$), the variance magnitude is therefore primarily determined by the weight scale. This observation motivates the use of training algorithms that explicitly control weight magnitudes to minimize $\sigma^2_w$, enabling variation-aware model optimization.

\subsection{Generalized Variance Modeling}
\label{sec:generalized-var-modeling}
The variance models for $\sigma^2_{G,\text{single}}$, $\sigma^2_{\Delta G}$, and $\sigma^2_w$ derived in the previous sections are based on measurements at a fixed time point ($t = 20$ hr)~\cite{didin_characterization_2026}. However, ReRAM devices exhibit time-dependent variability due to retention effects, requiring a generalized formulation that captures temporal evolution.

To incorporate time dependence, we model the variance scaling factor as a function of time. Prior works have shown that the variance exhibit a linear relationship with $\log(t)$~\cite{chang2025error}. Based on this observation, we extend the prefactor $b_w$ in Equation~\ref{eqn:sigma-w} to a time-dependent form:
\begin{align}
    b_w(t) = b_{w,\text{rate}} \cdot \left[\log(t) + t_0\right],
    \label{eqn:bt}
\end{align}
where $b_{w,\text{rate}}$ denotes the growth rate and $t_0$ is a constant offset.

The parameter $t_0$ can be determined independently of $b_{w,\text{rate}}$ using two time-point measurements:
\[
t_0 = \frac{\log(t_1) - c \cdot \log(t_2)}{c - 1}, \quad\text{with }
c = \frac{b(t_2)}{b(t_1)}.
\]

Substituting $b_w(t)$ into Equation~\ref{eqn:sigma-w}, the variance in the weight domain can be generalized as:
\begin{align}
\begin{split}
    \sigma_w^2(w, t)
    &= b_w(t) \cdot \exp(a_w \cdot w) \\
    &= b_{w,\text{rate}} \cdot \left[\log(t) + t_0\right] \cdot \exp(a_w \cdot w).
    \label{eqn:sigma-w-t}
\end{split}
\end{align}

This formulation preserves the exponential dependence on $w$ derived earlier, while introducing a time-dependent scaling factor that captures retention-induced variability. 

In practical deployments, the target inference time $t$ directly affects model reliability and determines the optimal trade-off between accuracy and robustness. In this work, we focus on $t = 20$ hr and use the shorthand $b = b(t = 20\text{ hr})$ unless otherwise specified. Nevertheless, the proposed ReRAM-aware training framework is general and can be adapted to arbitrary $t$. The impact of $t$ on the optimal regularization coefficient $\lambda$ is further discussed in the Results section.

\begin{table*}[!tb]
\centering
\renewcommand{\arraystretch}{1.5} 
\begin{tabular}{|c|cc|cc|cccc|c|}
\hline
 & \multicolumn{2}{c|}{$\sigma_{G,single}^2$} & \multicolumn{2}{c|}{$\sigma_{\Delta G}^2$} & \multicolumn{4}{c|}{$\sigma_{w}^2$} &  \\ \hline
Symbols & $a_{cell}$ & $b_{cell}$ & $a_{\Delta G}$ & $b_{\Delta G}$ & $\|w\|$ & $r_{G2w}$ & $a_w$ & $b_w$ & $a_{\Delta G}'$ \\
Formula & fitted & fitted & $-a_{cell}$ & $b_{cell}\cdot e^{a\cdot G_{max}}$ & from model & $\|w\|/(\Delta_g)_{max}$ & $a_{\Delta G}\times r_{G2w}^{-1}$ & $b_{\Delta G}\times r_{G2w}^2$ & $a_{\Delta G}\times (\Delta_g)_{max}$ \\
Values & -0.0593 & 26 & 0.0593 & 0.266 & 0.385 & 0.0064 & 9.244 & 1.095$\times 10^{-5}$ & 3.558 \\
Units & $\mu$S$^{-1}$ & $\mu$S$^2$ & $\mu$S$^{-1}$ & $\mu$S$^2$ & 1 & $\mu$S$^{-1}$ & 1 & 1 & 1 \\ \hline
\end{tabular}
\caption{Coefficients to model $\sigma_{G,single}^2$, $\sigma_{\Delta G}^2$, $\sigma_{w}^2$}
\label{tab:coeff}
\end{table*}

\subsection{ReRAM-implemented modules}

Although ReRAM arrays offer lower energy consumption, their crossbar structure is primarily optimized for matrix multiplication. Therefore, during deep learning inference, only matrix-based operations are mapped to ReRAM, while the remaining operations are executed on general-purpose hardware (e.g., CPU or digital accelerators). Fortunately, these matrix-based operations dominate both computation and memory usage in typical deep learning workloads.

\begin{figure}
    \centering
    \includegraphics[width=\linewidth]{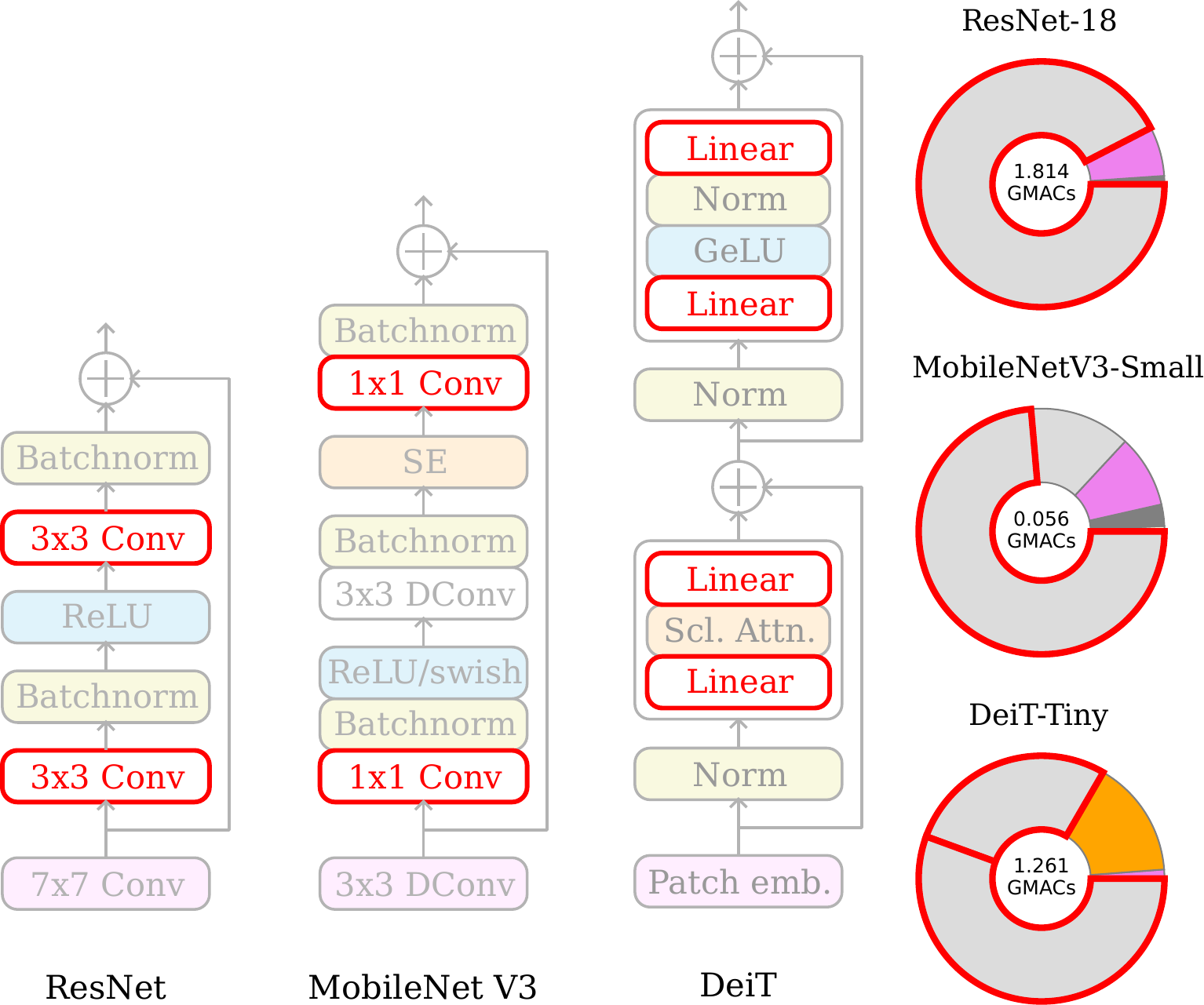}
    \caption{Modules implemented in ReRAM in different network families and their share in terms of MAC operations.}
    \label{fig:3models-rconv}
\end{figure}

Figure~\ref{fig:3models-rconv} illustrates this mapping strategy across three representative models: ResNet-18, MobileNetV3-Small, and DeiT-Tiny. Specifically, we implement 3$\times$3 convolutions and 1$\times$1 pointwise convolutions in ResNet-18 and MobileNetV3-Small, respectively, and linear layers in DeiT-Tiny using ReRAM. These operations account for approximately 92\%, 74\%, and 83\% of the total MAC operations in each model. This high coverage indicates that the energy efficiency benefits of ReRAM can effectively translate to system-level performance improvements.

\section{ReRAM-robust Model Training}

To enable robust inference on the ReRAM architecture described above, the device-level non-idealities characterized in Sections~\ref{SubSec:Diff_ReRAM}$\sim$\ref{sec:var-modeling} must be incorporated into training. We focus on two dominant sources of error: (i) input-dependent nonlinearity arising from the $\sinh$ I--V characteristics of ReRAM cells, and (ii) retention-induced weight noise, modeled through the variance formulation in Equation~\ref{eqn:sigma-w-t}.

Figure~\ref{fig:overview}(d--e) illustrates the \chingyi{training} framework. Starting from a pretrained model, we inject the input nonlinearity \chingyi{(Sec. \ref{sec:reram-cell-behavior}) with range-shrunk $sinh$ }into the forward pass to obtain a hardware-aware initialization. During the finetuning, \chingyi{in additional to optimize accuracy from this initialization,} retention-induced variability is incorporated as a variance-aware regularization term derived from $\sigma_w^2(w,t)$, penalizing weights according to their contribution to ReRAM-induced noise.

The final objective jointly optimizes the task loss and the variance-aware regularization, enabling the model to adapt to both nonlinear transformation and stochastic weight perturbations, and thereby improving robustness under realistic ReRAM deployment.

\begin{figure}[bt]
    \centering
    \includegraphics[width=\linewidth]{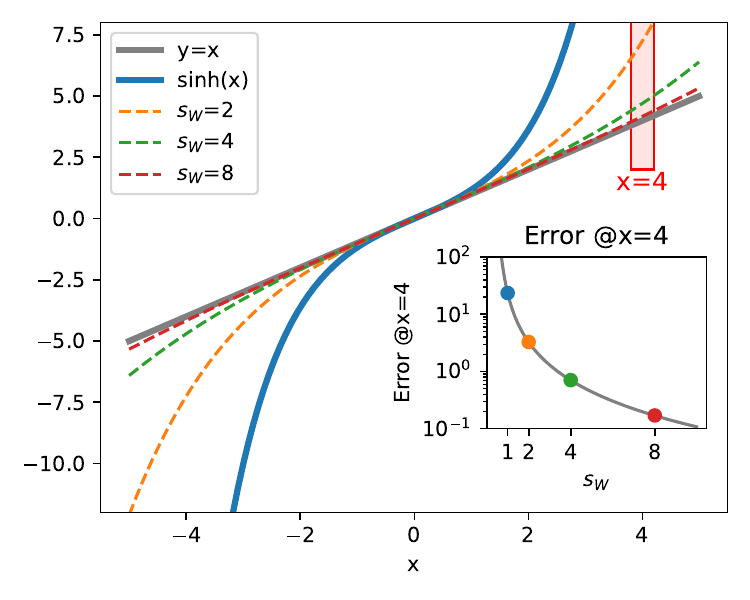}
    \caption{Range-shrunk sinh motivation: (a) Comparing $y=x$, $y=sinh(x)$, and $y=f_{s_W}(x)=s_W\times sinh(x/s_W)$ with different $s_W$ (b) Error between $y=f_{s_W}(x)$ and $y=x$ with different $s_W$}
    \label{fig:scaled-sinh}
\end{figure}

\subsection{Reducing Input Nonlinearity via Range-Shrunk \texorpdfstring{$\sinh$}{sinh}}
\label{sec:range-shrunk-sinh}

The $\sinh$-based transformation at the input is a key non-ideality in ReRAM crossbar computation. As shown in Figure~\ref{fig:scaled-sinh}(a), the output current of a ReRAM cell follows a $\sinh$ relationship with respect to the input voltage, rather than the ideal linear mapping ($y=x$). The discrepancy between the ideal Ohmic response and the nonlinear $\sinh$ transformation propagates across layers and accumulates at the model output, leading to accuracy degradation.

To mitigate this effect \chingyi{and build a hardware-aware initialization}, we adopt a range-shrunk $\sinh$ function defined as $f_{s_{W}}(x)= s_W \cdot \sinh(x / s_W)$, which improves linearity within a controlled input range\chingyi{ using a parameter $s_W$}. Figure~\ref{fig:scaled-sinh}(a) compares $f_{s_{W}}(x)$ for different values of $s_W=2,4,8$ against the ideal linear function and the baseline $\sinh(x)$. As $s_W$ increases, $f_{s_{W}}(x)$ becomes more linear, approaching $y=x$. This behavior is quantified in Figure~\ref{fig:scaled-sinh}(b), where the approximation error at a fixed input ($x=4$) decreases rapidly with increasing $s_W$. Intuitively, this \chingyi{can be explained} from $\sinh(x/s_W)/(x/s_W) \rightarrow 1$ as $x/s_W \rightarrow 0$.

\begin{figure}
    \centering
    \includegraphics[width=\linewidth]{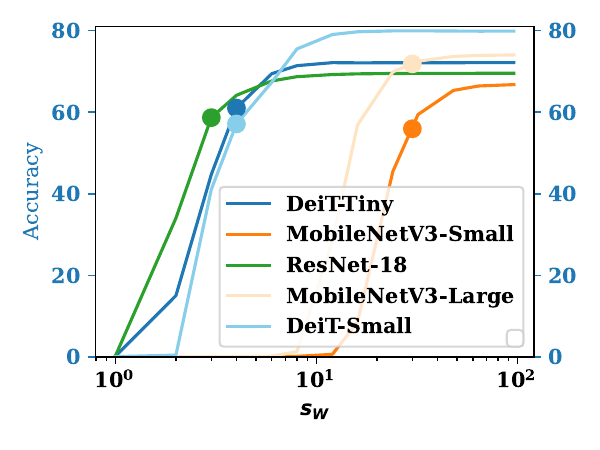}
    \caption{Accuracy-vs-$s_W$ for various models. The $s_W$ used is annotated with the corresponding colors}
    \label{fig:reram-r-vs-acc}
\end{figure}

The choice of $s_W$ directly determines the trade-off between nonlinearity error and model robustness, as it controls the effective input range. The optimal value of $s_W$ is model-dependent due to variations in input distributions $P(x)$. Figure~\ref{fig:reram-r-vs-acc} shows the accuracy as a function of $s_W$ across multiple models. As $s_W$ increases, the accuracy gradually recovers and saturates toward the corresponding digital baseline, reflecting reduced nonlinearity. Notably, different architectures exhibit different sensitivity; for example, MobileNetV3-Small requires a larger $s_W$ to reach saturation due to its distinct activation distribution.

In this work, we incorporate the range-shrunk $\sinh$ transformation into pretrained models to emulate ReRAM-induced nonlinearity, followed by finetuning to recover accuracy. A smaller $s_W$ introduces stronger distortion and increases finetuning difficulty, while an excessively large $s_W$ yields near-ideal initialization but compresses the effective input range, making the system more susceptible to noise. In practice, we select $s_W$ to induce approximately 10\% accuracy degradation prior to finetuning (as annotated in Figure~\ref{fig:reram-r-vs-acc}). Although selecting $s_W$ requires evaluating validation accuracy, we find that a simple binary search over the logarithmic range $s_W \in [1, 64]$ converges within 3--4 iterations.

\subsection{ReRAM Variance Minimization through Exponential Regularization}
\label{sec:exp-reg}

In addition to minimizing the data loss $\mathcal{L}_{data}$, training a ReRAM-robust model can be formulated as jointly minimizing the variance of the matrix multiplication output $y_i=\Sigma_jw_{ij}x_j$. Assuming the inputs of each row $x_j$ are drawn from the same random variable $\mathbf{X}$, the output can be approximated as $y_i=x\times\Sigma_jw_{ij}$. Extending it to matrix level, given the input $x$, the sum of output variance can be shown as
\[
\sigma_Y^2 = \sum_i\sigma^2_{y,i}=\sum_i\sum_jx^2\sigma_{w,i,j}^2=x^2\times\sum_i\sum_j\sigma_{w,i,j}^2
\]
With $\sigma_Y^2$ proportional to $\sum_i\sum_j\sigma_{w,i,j}^2$, this indicates that minimizing the total weight variance $\sum_i\sum_j\sigma_{w,i,j}^2$ directly reduces the output variance.

Based on the exponential variance model in Equation~\ref{eqn:sigma-w}, we incorporate the variance into training as a regularization term. The resulting objective becomes
\begin{align}
\begin{split}
    \mathcal{L}
    &= \mathcal{L}_{data} + \lambda \sum_i\sum_j \sigma_{w,i,j}^2 \\
    &= \mathcal{L}_{data} + \lambda \, b_w \sum_i\sum_j\exp(a_w w_{ij}) \\
    &= \mathcal{L}_{data} + \lambda' \sum_i\sum_j \exp(a w_{ij}),
    \label{eqn:exp-regularization}
\end{split}
\end{align}
where $\lambda' = \lambda \cdot b_w$ absorbs constant factors, and $a = a_w$. This yields a simple exponential regularization that penalizes large-magnitude weights according to the device-aware variance model.

While this regularization effectively reduces ReRAM-induced variance, it introduces potential optimization challenges. In particular, the gradient of the exponential term scales \chingyi{exponentially} as $\exp(a w_{ij})$, which can lead to disproportionately large updates for certain weights. \chingyi{Specifically, d}uring backpropagation that $w_{ij} \leftarrow w_{ij} - \eta\times{\partial \mathcal{L}}/{\partial w_{ij}}$, the update rule becomes
\[
w_{ij}\leftarrow w_{ij} - \eta' \exp(a\times w_{ij}),
\]
where $\eta'=\eta\times a\lambda'/r_{G2w}$ is an effective learning rate. Due to the exponential dependence, large-amplitude weights may cause unstable updates or even sign flips in the weights when the learning rate is not sufficiently small.

This behavior is analogous to the instability observed in L1 regularization~\cite{beck2009fast}, whereas L2 regularization provides smoother and more stable updates. In our setting, this issue is mitigated by adopting a finetuning strategy with a relatively small learning rate, which ensures stable optimization while preserving the benefits of variance-aware regularization.

\section{Experimental Setup}

\subsection{Tasks and Models}

To evaluate the effectiveness of the proposed ReRAM-aware training algorithms, we consider both computer vision and natural language processing tasks. Specifically, we adopt ImageNet~\cite{5206848} and SQuAD v2~\cite{rajpurkar2018know}, following the benchmark practices in MLCommons~\cite{reddi2020mlperf}.

For models, targeting ReRAM-oriented edge inference, we select MobileNetV3~\cite{howard2019searching}, ResNet-18~\cite{he2016deep}, and DeiT~\cite{touvron2021training} for image tasks, and MobileBERT~\cite{sun2020mobilebert} and TinyBERT~\cite{jiao2020tinybert} for language tasks. This selection covers both convolutional and transformer-based architectures across vision and language modalities.

\subsection{Model Training Details}

We adopt a finetuning-based training strategy to ensure practicality and generalizability \chingyi{following modern trend of large-scale pretraining}. Similar to standard transfer learning approaches~\cite{howard2018universal}, each model is initialized from pretrained weights and finetuned with reduced training cost. For image tasks, we train all models for 50 epochs using the initial learning rates reported in their original works and a cosine learning rate scheduler. For language tasks, since the model size is significantly larger than the downstream tasks, we follow their documented finetuning setting\chingyi{, such as learning rate and data preprocessing,} and train for 500 steps with batch size 32.

Across all experiments, we set the regularization coefficient to $\lambda = 0.006$, and do not apply additional weight decay. All other training configurations follow the original implementations of each model.

\subsection{Variation-Free and ReRAM Scenarios}

To evaluate both accuracy and robustness under ReRAM deployment, we consider two testing scenarios: a variation-free scenario and a ReRAM deployment scenario. Both scenarios include the I-V nonlinearity modeled in Section~\ref{sec:var-modeling}, but differ in the presence of retention-induced noise.

In the variation-free scenario, model accuracy is evaluated without injecting weight noise. In contrast, the ReRAM deployment scenario incorporates stochastic weight perturbations by sampling the retention-induced noise 16 times based on the variance model in Section~\ref{sec:var-modeling}. The final performance is reported by averaging accuracy across these sampled runs.

\begin{table}[b]
\centering
\begin{tabular}{@{}cccc@{}}
\toprule
\midrule
\multirow{2}{*}{Model} & \multicolumn{3}{c}{ImageNet Accuracy} \\ \cmidrule(l){2-4} 
 & \chingyi{Digital} & Variation-free & ReRAM(@20hr) \\ \midrule
ResNet18 & 69.76 & 70.28 & 70.12 (+0.36) \\
MobileNetV3-small & 67.21 & 66.50 & 65.64 (-1.57) \\
MobileNetV3-large & 74.28 & 74.01 & 73.89 (-0.39) \\
DeiT-Tiny & 72.20 & 72.86 & 72.87 (+0.67) \\ \midrule
Mobile BERT & 78.84 & 78.44 & 78.25 (-0.59) \\
Tiny BERT & 77.09 & 76.57 & 76.60 (-0.49) \\ \midrule
\multirow{2}{*}{Model} & Pretrained & Variation-free & ReRAM(@20hr) \\ \cmidrule(l){2-4} 
 & \multicolumn{3}{c}{SQuAD v2 F-1} \\ \midrule \bottomrule
\end{tabular}
\caption{Model Accuracy in ImageNet and SQuAD v2}
\label{tab:acc-all}
\end{table}

\begin{figure}[!bt]
\centerline{\includegraphics[width=\columnwidth]{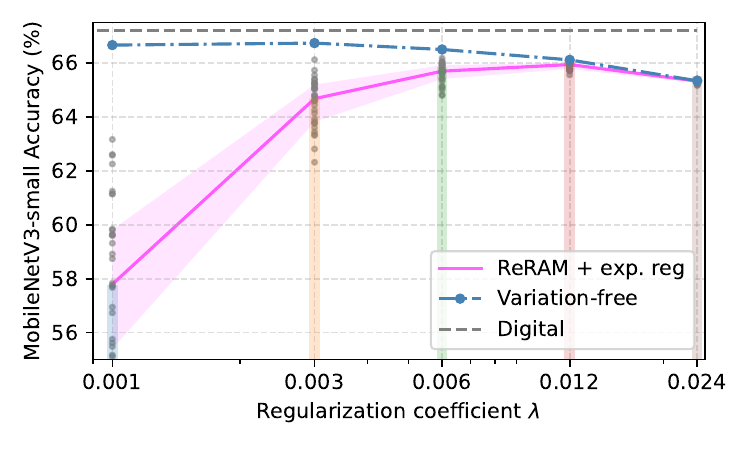}}
\caption{MobileNetV3-small Accuracy vs. $\lambda$ in variation-free and ReRAM scenarios. Every evaluation in ReRAM scenario is recorded in gray scatters, with mean as a solid line and 1st/3rd quartile in shaded region.}
\label{fig:acc}
\end{figure}
\section{Results}

\subsection{Accuracy in Variation-Free and ReRAM Scenarios}

Table~\ref{tab:acc-all} summarizes the accuracy of different models under variation-free and ReRAM deployment scenarios with non-idealities. For image classification \chingyi{tasks}, ResNet-18 and DeiT-Tiny exhibit negligible degradation under ReRAM variation compared to their digital counterparts. In contrast, the MobileNetV3 family shows more noticeable degradation, with MobileNetV3-Small experiencing the largest drop. We attribute this to the smaller network capacity of lightweight models, which makes them more sensitive to device-level noise. Conversely, larger models such as ResNet and DeiT benefit from higher model capacity and redundancy, which helps mitigate the impact of ReRAM-induced variation.

For language tasks, both MobileBERT and TinyBERT exhibit a degradation of approximately 0.5--1 F1 score on the SQuAD v2 dataset. \chingyi{Although the evaluation metric differs from that used in image classification tasks, the small degradation can be similarly explained through the more parameters in these language models (14.5M for Tiny BERT, 24.6M for Mobile BERT, and 11.7M for ResNet-18), consistent with the trend we observed in the vision models.}

\subsection{Regularization Effect with Different $\lambda$}

Retention-induced errors are mitigated in our framework by incorporating variance-aware regularization during finetuning. Similar to classical regularization, increasing the regularization coefficient $\lambda$ reduces model complexity but may degrade accuracy, reflecting a bias--variance trade-off~\cite{krogh1991simple}.

Figure~\ref{fig:acc} illustrates this trade-off by sweeping $\lambda$. As $\lambda$ increases, the variation-free accuracy decreases due to reduced emphasis on $\mathcal{L}_{data}$ in Equation~\ref{eqn:exp-regularization}. At the same time, the performance gap between variation-free and ReRAM scenarios shrinks, as stronger regularization reduces the variance of the trained model.

Notably, variation-free accuracy serves as an upper bound for ReRAM accuracy. Smaller $\lambda$ yields higher variation-free accuracy but suffers from larger degradation under ReRAM noise, whereas larger $\lambda$ improves robustness but lowers the achievable upper bound. This trade-off results in a concave trend in ReRAM accuracy, suggesting the existence of an optimal $\lambda$. In practice, we observe that $\lambda > 0.01$ is suboptimal, as the reduction in variation-free accuracy outweighs the gains in robustness.

\subsection{Comparison with L2 Regularization}

The proposed exponential regularization is designed to minimize both the task loss $\mathcal{L}_{data}$ and the ReRAM-induced variance during finetuning. While L2 regularization also penalizes large weights~\cite{hoerl1970ridge}, it does not explicitly account for the device-level noise characteristics derived in this work.

\begin{figure}[b]
\centerline{\includegraphics[width=\columnwidth]{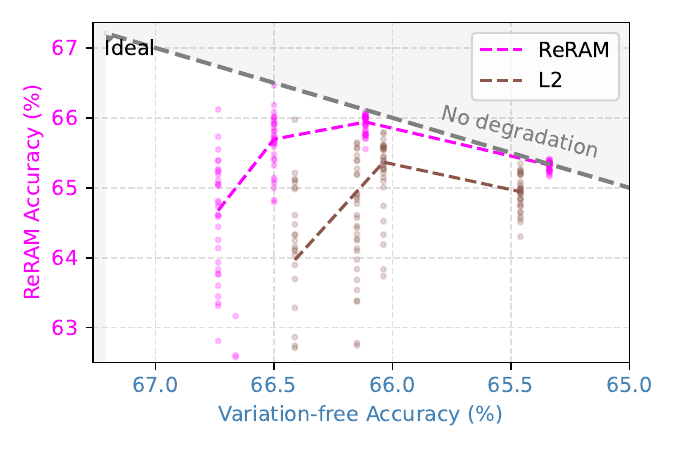}}
\caption{Comparison between exponential and L2 regularization in MobileNetV3-small. Two regularization methods are evaluated in both variation-free and ReRAM-scenario.}
\label{fig:acc-vs-l2}
\end{figure}

Figure~\ref{fig:acc-vs-l2} compares exponential and L2 regularization on MobileNetV3-Small. Each point in the figure corresponds to a trained model with a specific regularization strength. The x-axis shows the variation-free accuracy (i.e., accuracy without injected noise), which serves as an upper bound, while the y-axis shows the accuracy under the ReRAM deployment scenario with noise.

This plot highlights the trade-off between nominal accuracy and robustness. A better training method should achieve higher ReRAM accuracy for the same variation-free accuracy, forming a superior Pareto front in this space.

As shown in Figure~\ref{fig:acc-vs-l2}, models trained with exponential regularization consistently lie above those trained with L2 regularization, indicating higher ReRAM accuracy at comparable variation-free accuracy. This demonstrates that aligning the regularization term with the exponential variance model leads to more effective robustness optimization compared to generic L2 regularization.

\begin{figure}[!bt]
\centerline{\includegraphics[width=\columnwidth]{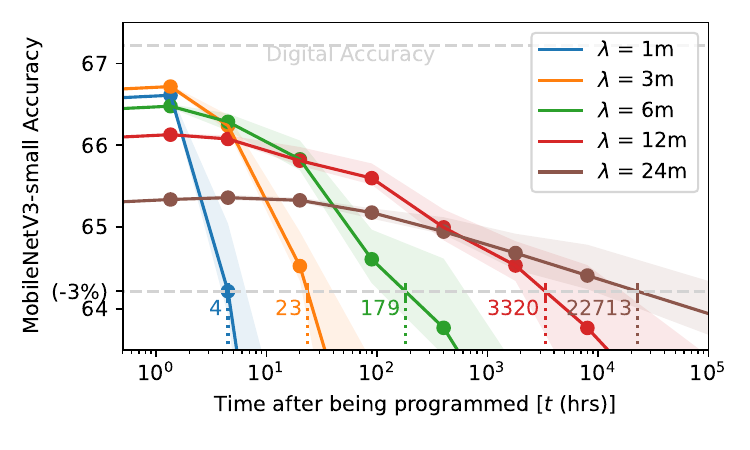}}
\caption{Accuracy vs. time after being programmed. The time taken with 3\% degradation is color-coded and labeled.}
\label{fig:acc-t}
\end{figure}

\subsection{Comparison with Related Hardware-aware Training Works}

\begin{figure}[!t]
\centerline{\includegraphics[width=\columnwidth]{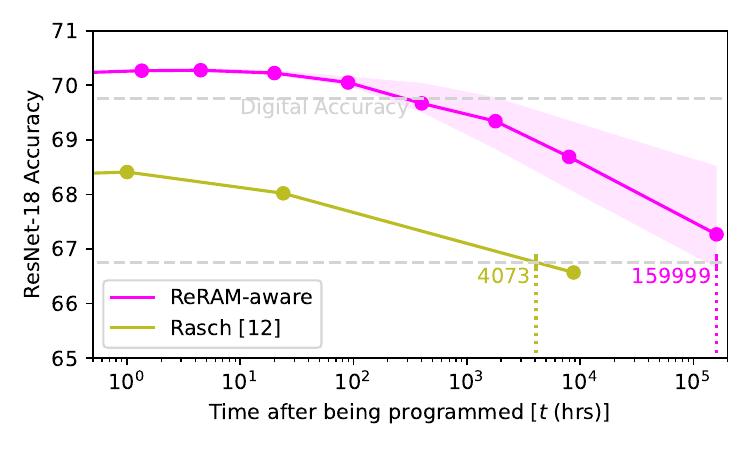}}
\caption{Comparison between proposed ReRAM-aware training and hardware-aware training including dynamic-range limitations, weight-programming errors, PCM drift and system noise~\cite{rasch2023hardware}}
\label{fig:acc-vs-ibm}
\end{figure}

This work accounts for ReRAM non-idealities, specifically a static non-linear I-V relationship and dynamic retention-induced errors that cause temporal accuracy degradation. To validate our simulation framework, we compare the temporal accuracy trend of our ReRAM deployment against a hardware-aware training work~\cite{rasch2023hardware}. Figure \ref{fig:acc-vs-ibm} illustrates the ResNet-18 accuracy with weight variation $\sigma_w^2(w,t)$ modeled according to Section~\ref{sec:generalized-var-modeling}. For benchmarking, the accuracy of another hardware-training work~\cite{rasch2023hardware} after 1 second, 1 hour, 1 day, and 1 year is plotted as an olive-colored curve. Our results exhibit a degradation trend along $log(t)$ that is consistent with the prior work, albeit with a vertical offset in absolute accuracy. It is important to note that this comparison does not aim to claim superior accuracy, as direct performance comparisons are often confounded by differing hardware deployment parameters. The objective of Figure \ref{fig:acc-vs-ibm} is to validate the physical realism of our experimental setup regarding time-varying retention-induced errors.

\subsection{Accuracy over Varying Retention Time}

Retention-induced errors increase over time after programming, making ReRAM accuracy inherently time-dependent. Figure~\ref{fig:acc-t} shows the accuracy of MobileNetV3-Small as a function of time for different regularization coefficients $\lambda$. This can be viewed as a generalized extension of Figure~\ref{fig:acc}, where the variation-free and ReRAM deployment scenarios correspond to $t=0$ and $t=20$ hr, respectively.

Beyond these two time points, Figure~\ref{fig:acc-t} provides a comprehensive view of model robustness over time. We quantify robustness using the retention time required to maintain accuracy above a threshold (e.g., 64\%, corresponding to a 3\% drop from digital accuracy). This threshold is indicated by the horizontal line $y = acc_{\text{digital}} - 3$.

The time to reach this degradation varies significantly across models: from approximately 5 hours for weak regularization ($\lambda=0.001$) to over 30 months for strong regularization ($\lambda=0.024$). This highlights a clear trade-off between initial accuracy and long-term robustness, which is critical for real-world ReRAM deployments.


\begin{figure}[!t]
\centerline{\includegraphics[width=\columnwidth]{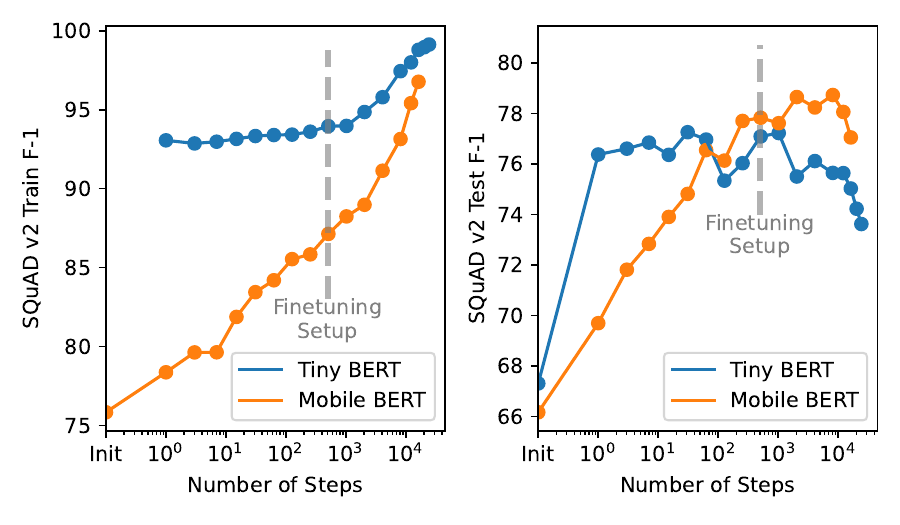}}
\caption{Accuracy vs. number of steps for TinyBERT and MobileBERT in SQuAD v2 dataset. The steadily increased training accuracy and the concave test accuracy trend show that the model starts overfitting after 500 steps.}
\label{fig:bertacc-vs-step}
\end{figure}

\subsection{Language Model Finetuning via Early Stopping}

In this study, we adapt large-scale pre-trained models to small downstream tasks through finetuning. Given the significant difference between the model capacity and the downstream dataset size, the finetuning process suffers from catastrophic forgetting of upstream knowledge and overfitting to the downstream task. These phenomena are characterized by a simultaneous decline in upstream task performance and downstream test accuracy, even as the downstream training accuracy continues to improve.

To mitigate overfitting effect, we employ early-stopping to restrict the number of training steps. Figure \ref{fig:bertacc-vs-step} illustrates the accuracy trajectories for TinyBERT and MobileBERT across varying number of steps, with \"Init\" the model after applying range-shrunk $sinh$ in Section \ref{sec:range-shrunk-sinh}. While the training accuracy increases monotonically with additional steps, the test accuracy reaches a peak mid-training before degradation due to overfitting. Consequently, to ensure robust generalization across our language model fine-tuning tasks, we fixed the training duration at 500 steps. The covering training size ($500\times 32$) is eight times smaller than the training set size $130$k

\section{Conclusion}

This work addresses the critical challenge of ReRAM non-idealities in deploying deep neural networks on in-memory computing architectures. To mitigate accuracy degradation, we propose a finetuning-based training framework that minimizes ReRAM-induced variance while restoring model performance with significantly lower overhead than training-from-scratch approaches.

Our method targets two key non-idealities: I--V nonlinearity and retention-induced variability. Specifically, we apply a range-shrunk $\sinh$ transformation to reduce input nonlinearity, and incorporate a variance-aware regularization term derived from the device-level noise model to account for retention effects during optimization.

Experimental results across diverse tasks, including image classification and question answering, demonstrate the effectiveness of the proposed approach. We achieve less than 2\% accuracy degradation on ImageNet and less than 1 F1 score degradation on SQuAD v2 under ReRAM deployment. 

Overall, this finetuning framework provides a practical and efficient solution for improving robustness in analog AI systems, aligning well with modern model adaptation pipelines and enabling scalable deployment on ReRAM-based hardware.

\section*{Acknowledgment }
This work was partly supported by both Semiconductor Research Corporation (Award \#2023-AM-3160.032). The authors acknowledge the University of Maryland supercomputing resources (http://hpcc.umd.edu) made available for conducting the research reported in this paper.

\section*{References}

\bibliographystyle{IEEEtran}
\bibliography{references.bib,references_shah}

\newpage
\newpage




\newpage
\section*{Appendix}

\begin{figure}[!t]
\centerline{\includegraphics[width=\columnwidth]{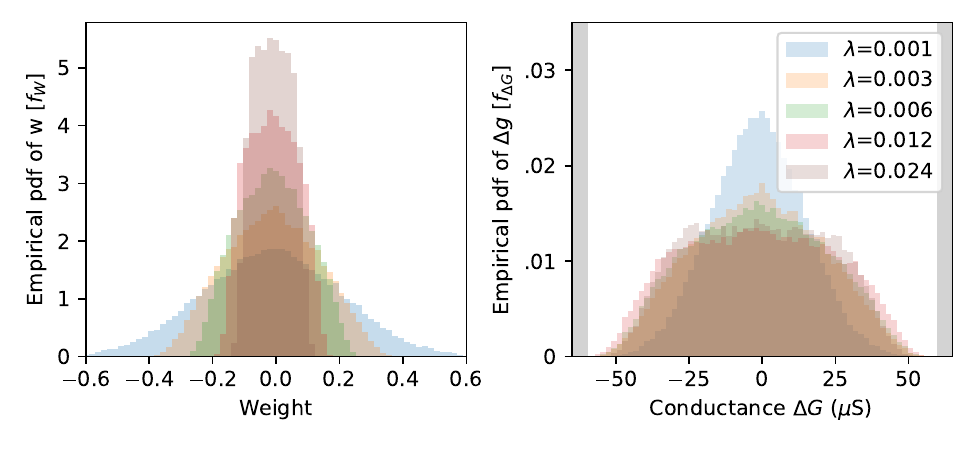}}
\caption{Probability density function of (a) weight $f_W(w)$ and (b) conductance difference $f_{\Delta G}(\Delta G)$ in MobileNetV3-small trained with different $\lambda$}
\label{fig:pw-pg}
\end{figure}

\section{Empirical Weight and Conductance Distribution}

In this work, the regularization coefficient $\lambda$ controls the trade-off between minimizing ReRAM-induced variance and preserving variation-free accuracy. From a high-level perspective, larger $\lambda$ imposes a stronger penalty on weights during training (Section~\ref{sec:exp-reg}), discouraging large-magnitude weights that contribute disproportionately to ReRAM variance.

Figure~\ref{fig:pw-pg}(a) illustrates this effect through the empirical probability density function (p.d.f.) of weights in a representative layer (features.11.conv.0.conv.weight). As $\lambda$ increases, the weight distribution becomes increasingly concentrated around zero, with a reduced spread and smaller maximum magnitude $\|w\|_{\max}$. Conversely, smaller $\lambda$ allows a wider distribution with heavier tails, indicating the presence of larger weights.

Through the conductance mapping in Equations~\ref{eqn:conductance-mapping1} and~\ref{eqn:conductance-mapping2}, the conductance difference between a pair of ReRAM cells representing a weight $w$ can be expressed as $\Delta G = w / r_{G2w}$, where $r_{G2w} = \|w\|_{\max} / (G_{\max} - G_{\min})$. Substituting this relation yields
\[
\Delta G = \frac{w}{\|w\|_{\max}} \cdot (G_{\max} - G_{\min}),
\]
which effectively normalizes the weight distribution by its maximum magnitude.

Figure~\ref{fig:pw-pg}(b) shows the resulting p.d.f. of $\Delta G$. Due to the $\|w\|_{max}$-based normalization, a different trend emerges compared to the weight distribution. While larger $\lambda$ reduces the absolute scale of weights, it also compresses $\|w\|_{\max}$, causing the normalized weights $w / \|w\|_{\max}$ to spread more evenly across the valid conductance range. As a result, the $\Delta G$ distribution becomes flatter and more uniform. In contrast, smaller $\lambda$ leads to a few dominant large weights, resulting in a more concentrated $\Delta G$ distribution.

These observations provide an intuitive interpretation of the role of $\lambda$: increasing $\lambda$ reduces the dynamic range of weights, thereby lowering weight variance $\sigma_W$, while simultaneously redistributing the normalized conductance differences more uniformly. This behavior explains the trade-off observed in the main text between accuracy and robustness under different regularization strengths.


\section{Weight-inferred ReRAM Variance}

\begin{figure}[!t]
\centerline{\includegraphics[width=\columnwidth]{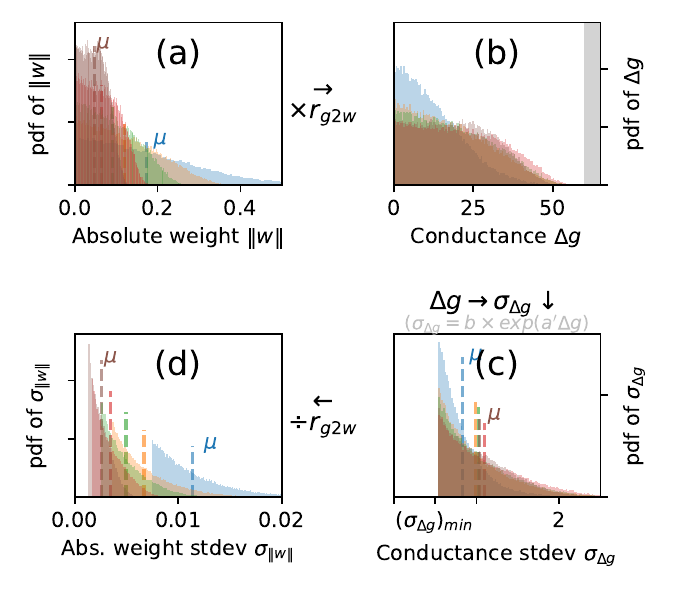}}
\caption{Probability density function $f$ of (a) absolute weight $\|w\|$, (b) pair conductance difference $\Delta G$, (c) standard deviation of pair conductance difference $\sigma_{\Delta G}$, and (d) standard deviation of absolute weight $\sigma_{\|w\|}$. These pdfs can be converted through $r_{G2w}$ scaling, variance modeling (Equation \ref{eqn:var-g}), $1/r_{G2w}$ scaling}
\label{fig:sigmaw}
\end{figure}

To analyze how ReRAM retention noise propagates into model parameters, Fig.~\ref{fig:sigmaw} derives the weight-inferred ReRAM variance $\sigma_w^2$ from the empirical weight distribution. The procedure consists of three steps. First, in Fig.~\ref{fig:sigmaw}(a)$\rightarrow$(b), the distribution of absolute weights $\lVert w \rVert$ is mapped to the conductance-difference domain using the scaling factor
\[
\frac{1}{r_{G2w}}=\frac{G_{\max}-G_{\min}}{\lVert w\rVert_{\max}}.
\]
Second, in Fig.~\ref{fig:sigmaw}(b)$\rightarrow$(c), the conductance distribution is converted to the conductance standard deviation $\sigma_{\Delta G}$ through the device-aware variance model in Eq.~\ref{eqn:var-g}. Finally, in Fig.~\ref{fig:sigmaw}(c)$\rightarrow$(d), the conductance-domain deviation is mapped back to the weight domain through the scaling factor $r_{G2w}$, yielding the corresponding weight standard deviation $\sigma_{\lVert w\rVert}$ and hence the weight-inferred variance $\sigma_w^2$.

The effect of $\lambda$ to $\sigma_{\|w\|}$ is most clearly illustrated by the mean values, marked by dashed lines in Fig.~\ref{fig:sigmaw}(d). Stronger regularization leads to a smaller weight-domain deviation and, correspondingly, less ReRAM-induced accuracy degradation. This reduction can be understood as the combined effect of two factors: the model-dependent weight range $\lVert w\rVert_{\max}$ and the conductance-domain distribution. Increasing $\lambda$ reduces $\lVert w\rVert_{\max}$ to shrink the range of $\|w\|$, while at the same time making the conductance distribution more uniform and slightly increasing $\sigma_{\Delta G}$. These two effects influence the weight-inferred ReRAM variance in opposite directions. However, Fig.~\ref{fig:sigmaw}(c) shows that the mean of $\sigma_{\Delta G}$ remains relatively close across different $\lambda$, indicating that the dominant reduction in $\sigma_{\lVert w\rVert}$ mainly comes from the shrinkage of $\lVert w\rVert_{\max}$. This trend is consistent with the positive correlation between $\sigma_{\lVert w\rVert}$ and $\lVert w\rVert_{\max}$, and further explains why larger $\lambda$ is effective during finetuning.

Another important observation is the difference between the lower bounds at the left of the distribution in Fig.~\ref{fig:sigmaw}(c) and Fig.~\ref{fig:sigmaw}(d). The lower bound in Fig.~\ref{fig:sigmaw}(c) is consistent across models because it is set by the minimum conductance deviation $\sigma_{\min,\Delta G}$ obtained from ReRAM measurements in Fig.~\ref{fig:reram-architecture}(d). By contrast, the lower bound in the weight domain is derived from
\[
\sigma_{\min,\lVert w\rVert}
=
\frac{\lVert w\rVert_{\max}}{G_{\max}-G_{\min}}\,
\sigma_{\min,\Delta G}
\propto
\lVert w\rVert_{\max},
\]
which depends directly on the model-specific weight range. By the same argument, the full weight-domain deviation interval $[\sigma_{\min,w},\,\sigma_{\max,w}]$ is primarily determined by $\lVert w\rVert_{\max}$, while the conductance distribution shapes how probability mass is distributed within that interval. Since Fig.~\ref{fig:sigmaw}(c) shows only modest variation in the $\sigma_{\Delta G}$ distribution across $\lambda$, the range-setting factor $\lVert w\rVert_{\max}$ becomes the dominant quantity controlling the overall weight noise.

Overall, separating the contributions of $\sigma_{\min,\lVert w\rVert}$ and $\sigma_{\Delta G}$ clarifies that the improvement from stronger regularization arises mainly from suppressing the weight range rather than significantly altering the device-level conductance deviation. This observation further justifies the use of larger $\lambda$ to obtain smaller $\lVert w\rVert_{\max}$, even though the normalized conductance distribution becomes more uniform, as observed in Fig.~\ref{fig:pw-pg}.

\section{Priors in Exponential Regularization}

In this work, ReRAM-induced variance is reduced through exponential regularization, where the model parameters are optimized as
\begin{equation}
    \mathbf{w}^* = \arg \min_{\mathbf{w}} \left[ \sum_i \mathcal{L}_{\text{data},i} + \lambda \sum_j e^{a|w_j|} \right].
    \label{eqn:optimization}
\end{equation}

This objective can be interpreted from a probabilistic perspective as a maximum a posteriori (MAP) estimation problem. Specifically, Eq.~\ref{eqn:optimization} can be rewritten in exponential form as
\begin{align}
\begin{split}
    \mathbf{w}^*
    &= \arg \max_{\mathbf{w}} \exp\left(-\sum_i \mathcal{L}_{\text{data},i}\right)
       \times \exp\left(-\lambda \sum_j e^{a|w_j|}\right) \\
    &= \arg \max_{\mathbf{w}} \left[ \prod_i \exp(-\mathcal{L}_{\text{data},i}) 
       \times \prod_j \exp\left(-\lambda e^{a|w_j|}\right) \right]
    \label{eqn:exp-exp-prior}
\end{split}
\end{align}

Equation~\ref{eqn:exp-exp-prior} decomposes the posterior into a likelihood $P(data\mid\mathbf{w})$ and a prior $P(\mathbf{w})$:
\[
P(\mathbf{w} \mid \text{data}) \propto 
\underbrace{\prod_i \exp(-\mathcal{L}_{\text{data},i})}_{\text{likelihood}}
\times
\underbrace{\prod_j \exp\left(-\lambda e^{a|w_j|}\right)}_{\text{prior}}.
\]

Thus, the implied prior on each weight takes the form
\[
P(w_j) \propto \exp\left(-\lambda e^{a|w_j|}\right),
\]
which corresponds to a nested exponential (``exp-exp'') prior.

Compared to the Gaussian prior induced by L2 regularization, this prior exhibits a significantly sharper decay for large $|w_j|$, leading to stronger suppression of large-magnitude weights. This behavior is consistent with the exponential dependence of ReRAM variance on weight magnitude derived in Section~III, making the regularization term well-aligned with the underlying device noise characteristics.

In the following discussion, we leverage this exp-exp prior formulation to fit and analyze the empirical probability density functions observed in trained models.




\section{Fitting Probability Density Function of Weight $P(w)$}
\label{sec:fitting-weight-priors}

\begin{figure}[bt]
\centerline{\includegraphics[width=\columnwidth]{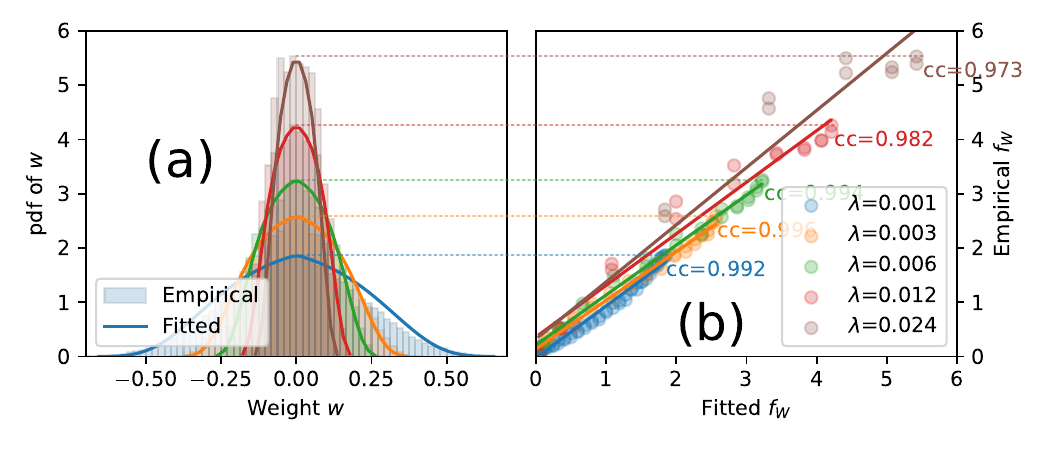}}
\caption{(a) Empirical and fitted p.d.f. $f_W$ versus $w$ (b) Scatter plot and correlation coefficients between empirical $f_W(w)$ and fitted $f_W(w)$}
\label{fig:cc-vs-w}
\end{figure}

\begin{figure}[bt]
\centerline{\includegraphics[width=\columnwidth]{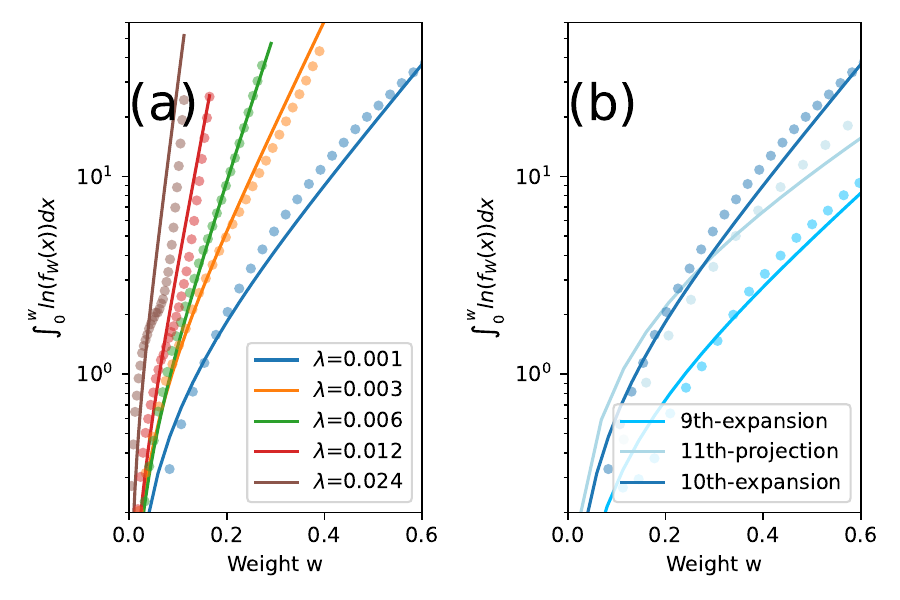}}
\caption{(a) $\int_0^w \ln(f_W(x)) dx$ versus weight $w$ in the pointwise convolution layer (10th layer) under different regularization coefficients $\lambda$, showing increasing slope with larger $\lambda$ (b) $\int_0^w \ln(f_W(x)) dx$ versus $w$ across different layers in a model trained with $\lambda=0.006$}
\label{fig:b-vs-w-div}
\end{figure}

Equation~\ref{eqn:exp-exp-prior} suggests that ReRAM-aware training induces an exp-exp prior over weights. To validate this hypothesis, we fit the empirical probability density function (p.d.f.) using a transformation that reduces the nested exponential form to a single-parameter representation.

Starting from the assumed p.d.f. with respect to weight
\begin{align}
\begin{split}
    f_W(w) &= s^{-\lambda \exp(a\|w\|)}, \\
    \ln\left(\frac{1}{f_W(w)}\right) &= \ln(s)\times\lambda\exp(a\|w\|),
\end{split}
\end{align}
we apply an integral transformation:
\begin{align}
\begin{split}
    \int_0^w \ln\left(\frac{1}{f_W(x)}\right) dx
    &= \int_0^w \ln(s)\times\lambda\,\exp(ax)\,dx \\
    &= \ln(s)\times\lambda\times\frac{1}{a}\left(e^{aw} - 1\right).
\end{split}
\label{eqn:pdf-fit-wexp}
\end{align}

This transformation converts the nested exponential form into a scaled exponential function. The scaling factor is given by $\ln(s)\lambda/a$, while the exponential term $e^{aw}-1$ captures the dependence on $w$.

Here, $a$ is determined from the device-aware variance model:
\[
a = \frac{a_{\Delta G}}{r_{G2w}} 
= a_{\Delta G} \cdot \frac{G_{\max}-G_{\min}}{\|w_{\max}\|},
\]
using known device parameters ($G_{\max}$, $G_{\min}$, $a_{\Delta G}$) and the model-dependent $\|w_{\max}\|$ (cf. Eq.~\ref{eqn:sigma-w}).

Based on Eq.~\ref{eqn:pdf-fit-wexp}, we define a new function $f'(w)$ as
\begin{align}
\begin{split}
    f'(w) &= \frac{1}{e^{aw}-1}\int_0^w \ln\left(\frac{1}{f_W(x)}\right)dx \\
    &= \ln(s)\,\lambda\,\frac{1}{a} = b_w,
\end{split}
\label{eqn:pdf-overexp}
\end{align}
Ideally, $f'(w)$ should be a constant independent of $w$ after canceling the exponential term. Therefore, this transformed function provides a convenient scalar quantity for characterizing the p.d.f.

In practice, $b_w$ is estimated as the median of $f'(w)$ over all positive weights. To validate this formulation, Fig.~\ref{fig:b-vs-w-div} compares the empirical $\int_0^w \ln(1/f_W(x)) dx$ with the fitted curve $b_w(e^{aw}-1)$, showing strong agreement. The visualization is performed in this form (instead of directly plotting $f'(w)$) to highlight the exponential trend implied by the model.

We further evaluate consistency across (i) different regularization strengths $\lambda$ within the same layer and (ii) different layers within the same model. In both cases, the transformed distributions exhibit consistent exponential behavior, confirming the validity of the exp-exp prior assumption.

Figure~\ref{fig:cc-vs-w}(a) provides a direct comparison between empirical and fitted p.d.f.s, while Fig.~\ref{fig:cc-vs-w}(b) quantifies the fitting accuracy through correlation analysis. The high correlation coefficients, ranging from 0.973 to 0.996, demonstrate that the proposed fitting procedure in Eq.~\ref{eqn:pdf-overexp} accurately captures the empirical weight distribution.
\section{Fitting Priors of Normalized Weights / Conductance}

In addition to fitting the distribution of model weights $w$, we are also interested in hardware-visible quantities such as conductance $G$ and normalized weights $\tilde{w}$. Our fitting framework can be directly extended to the p.d.f. of conductance $f_G(G)$ and its normalized forms. Here, we use normalized conductance $\tilde{G}$ as an example, while other variables can be obtained through standard p.d.f. scaling.

Based on the relation
\[
\tilde{G} = \frac{G}{G_{\max}} = \frac{w}{\|w\|_{\max}},
\]
the weight-domain p.d.f. in Eq.~\ref{eqn:pdf-fit-wexp} can be rewritten as
\begin{align}
\begin{split}
    f_{\tilde{G}}(\tilde{G})
    &= \|w\|_{\max} \times f_W(\tilde{G}\times\|w\|_{\max}) \\
    &= \|w\|_{\max} \times s^{-\lambda \exp\!\left(a|\tilde{G}|\times\|w\|_{\max}\right)} \\
    &= \|w\|_{\max} \times s^{-\lambda \exp\!\left(a_{\Delta G}(G_{\max}-G_{\min})\,|\tilde{G}|\right)}.
\end{split}
\label{eqn:fg}
\end{align}

In Eq. \ref{eqn:fg}, the exponent scale $a_{\Delta G}(G_{\max}-G_{\min})$ depends only on device parameters  in contrast to $f_W(w)$ with $a\propto\|w_{max}\|^{-1}$.

Following the same transformation in Eq.~\ref{eqn:pdf-fit-wexp}, we obtain
\begin{equation}
    \int_0^t \ln\left(\frac{1}{f_{\tilde{G}}(\tilde{G})}\right)\,d\tilde{G}
    =
    \ln(s)\,\lambda\,\frac{1}{a'_{\Delta G}}
    \left(e^{a'_{\Delta G} t} - 1\right).
\label{eqn:int-ln-fg}
\end{equation}

\begin{figure}[!t]
\centerline{\includegraphics[width=\columnwidth]{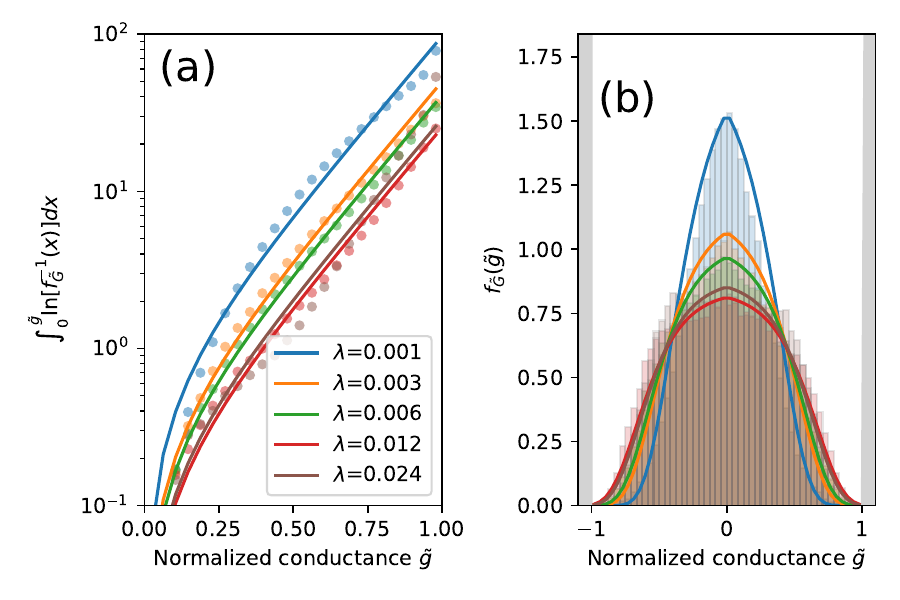}}
\caption{(a) $\int_0^t \ln(1/f_{\tilde{G}}(\tilde{G}))\,d\tilde{G}$ versus $\tilde{G}$ under different regularization coefficients $\lambda$. The consistent slope at large $\tilde{G}$ indicates a shared exponent $a'_{\Delta G}$ in Eq.~\ref{eqn:int-ln-fg}. (b) Empirical and fitted $P(\tilde{G})$ versus $\tilde{G}$ for different $\lambda$.}
\label{fig:b-vs-g-div}
\end{figure}

With $a'_{\Delta G} = a_{\Delta G}(G_{\max}-G_{\min})$, 
Eq. \ref{eqn:int-ln-fg} obtains a similar form as Eq. \ref{eqn:pdf-overexp}. While the functional form remains identical, an important distinction is that the exponential term $e^{a'_{\Delta G} t}-1$ is invariant across models due to the model-independent parameter $a'_{\Delta G}$. Figure~\ref{fig:b-vs-g-div}(a) visualizes conductance-domain Eq.~\ref{eqn:int-ln-fg} analogous to the weight-domain results in Fig.~\ref{fig:b-vs-w-div}. The model-independent $a'_{\Delta G}$ implies that models trained with different $\lambda$ differ only by a scaling factor, which appears as a vertical shift.

Another advantage of Eq. \ref{eqn:int-ln-fg} is that we can model the distribution using a single parameter $b=ln(s)\times\lambda/a'_{\Delta G}$. Fig. \ref{fig:b-vs-g-div}(b) shows both the empirical and fitted distribution. In the later section, we will use this property to discuss layer-level effect for regularization.

\section{Layer-wise Conductance Distribution}

While the previous sections establish that the learned parameters follow an exp-exp prior, we further investigate how the regularization coefficient $\lambda$ shapes the \emph{conductance-domain} distribution from layer to layer.

To visualize conductance distribution in layer granularity, we need to represent each layer in a visually simpler way. Figure~\ref{fig:b-vs-g-div} shows that the conductance p.d.f. becomes sharper as $\lambda$ decreases. To quantify this effect, we relate the distribution shape to the scaling factor in Eq.~\ref{eqn:int-ln-fg}. Since the transformed distribution is fully characterized by a single parameter, we interpret the constant $b$ in Eq.~\ref{eqn:pdf-overexp} as the \emph{sharpness} of the conductance p.d.f.

To validate the sharpness-modeled distribution, we calculate correlation between the empirical distribution and the one recovered from inversing Eq. \ref{eqn:fg}. To be more specific, given the sharpness $b$ and the predefined device-dependent parameter $a'_{\Delta G}$, the distribution can be recovered from
\begin{align}
\begin{split}
    f_{\tilde{G},\text{modeled}}(\tilde{G})
    &= \exp\!\left(-\frac{d}{d\tilde{G}}\,b\left(e^{a'_{\Delta G}\tilde{G}} - 1\right)\right) \\
    &= \exp\!\left(-b\,a'_{\Delta G}\,e^{a'_{\Delta G}\tilde{G}}\right).
\end{split}
\label{eqn:fg-modeled}
\end{align}

\begin{figure}[!t]
\centerline{\includegraphics[width=\columnwidth]{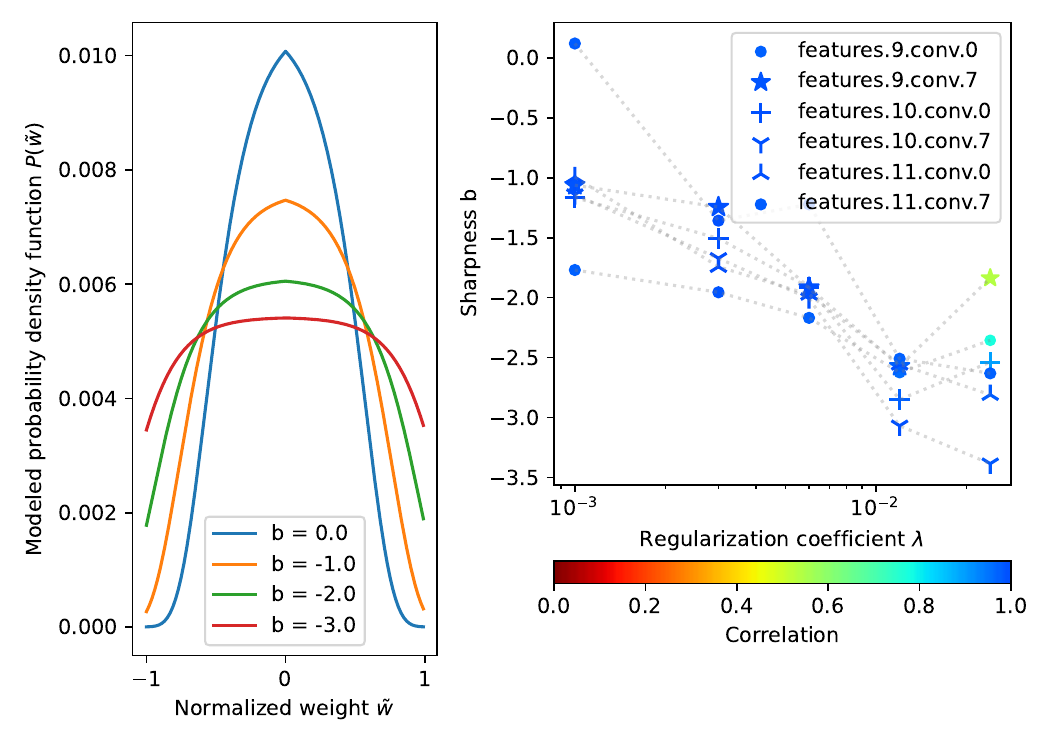}}
\caption{(a) Modeled $P(\tilde{G})$ for different sharpness values $b$. (b) Sharpness $b$ versus regularization coefficient $\lambda$ across layers, with color-coded correlation coefficients indicating fitting accuracy.}
\label{fig:b-vs-p}
\end{figure}

Eq. \ref{eqn:fg-modeled} is equivalent to Eq. \ref{eqn:int-ln-fg} through some algebratic manipulation. Figure~\ref{fig:b-vs-p}(a) visualizes $f_{\tilde{G},\text{modeled}}$ for different values of $b$, confirming that $b$ effectively controls the sharpness of the distribution: larger $b$ leads to a more concentrated distribution, while smaller $b$ results in a flatter one.

An important observation is that $b$ appears larger for smaller $\lambda$, which may seem contradictory to Eq.~\ref{eqn:pdf-overexp}, where $b = \ln(s)\lambda/a$ suggests proportionality to $\lambda$. However, this interpretation is incomplete because $\ln(s)$ is also dependent on $\lambda$. Therefore, $b$ should be regarded as an effective fitting parameter capturing the intercept of the transformed representation, rather than a quantity with direct physical proportionality to $\lambda$.

Figure~\ref{fig:b-vs-p}(b) extends this analysis across multiple ReRAM layers in MobileNetV3-Small. We observe a consistent negative correlation between $\lambda$ and $b$, indicating that stronger regularization produces less sharply concentrated conductance distributions. Additionally, the color-coded correlation coefficients show that for large $\lambda$, the agreement between empirical and modeled distributions slightly degrades, suggesting that overly strong regularization pushes the distribution away from the ideal exp-exp prior.

\begin{figure}[!t]
\centerline{\includegraphics[width=\columnwidth]{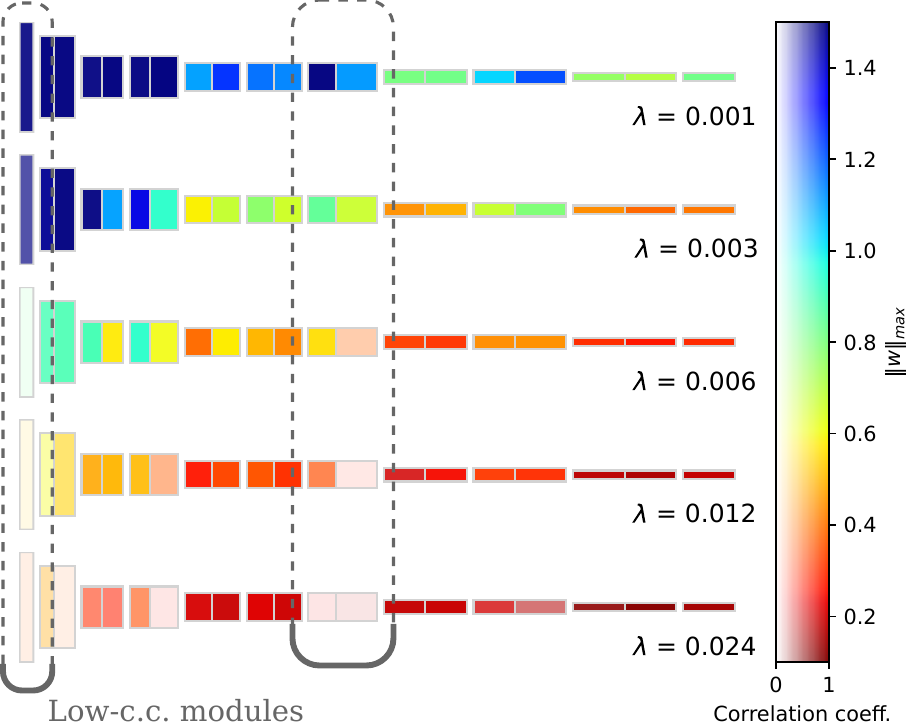}}
\caption{Per-layer $\|w\|_{\max}$ and correlation coefficient in MobileNetV3-Small trained with different $\lambda$. Stronger regularization reduces $\|w\|_{\max}$ while slightly degrading the exp-exp prior fitting in certain layers.}
\label{model-arch}
\end{figure}

To further analyze layer-wise behavior, Fig.~\ref{model-arch} visualizes $\|w\|_{\max}$ and the corresponding fitting correlation. Larger $\lambda$ consistently reduces $\|w\|_{\max}$, as indicated by warmer colors, confirming the relationship between regularization strength and weight range.

At the same time, we observe decreasing correlation coefficients for larger $\lambda$, particularly in certain layers (highlighted in gray). This indicates that strong regularization not only reduces the dynamic range but also alters the distribution shape, causing deviation from the ideal exp-exp prior.

\section{Fitting Priors without Known $a$}

The fitting procedure above assumes that the parameter $a_{\Delta G}$ (or $a'_{\Delta G}$) is known from device-level measurements. In practice, however, such parameters may not always be available for a given device. Therefore, we propose an alternative fitting method that estimates $a$ directly from empirical weight distributions.

With sufficiently large $w$, we can approximate Eq.~\ref{eqn:pdf-fit-wexp} as
\begin{align}
\begin{split}
    \int_0^w \ln\left(\frac{1}{f_W(x)}\right) dx 
    &= \ln(s)\lambda\frac{1}{a}(e^{aw} - 1) \\
    &\approx \ln(s)\lambda\frac{1}{a}e^{aw},
\end{split}
\end{align}

This scaled exponential can be converted to a straight line through logarithm
\begin{align}
    \ln\left(\int_0^w \ln\left(\frac{1}{f_W(x)}\right) dx\right)
    \approx aw + \left[\ln(\ln(s)) + \ln(\lambda) - \ln(a)\right],
    \label{eqn:ln-int-ln-fw}
\end{align}

This transformation (log-integral-log) converts the exp-exp distribution into a straight line with a slope $a$ and an intercept $ln(ln(s))+ln(\lambda)-ln(a)$, enabling direct estimation of $a$ from data.

\begin{figure}[!t]
\centerline{\includegraphics[width=\columnwidth]{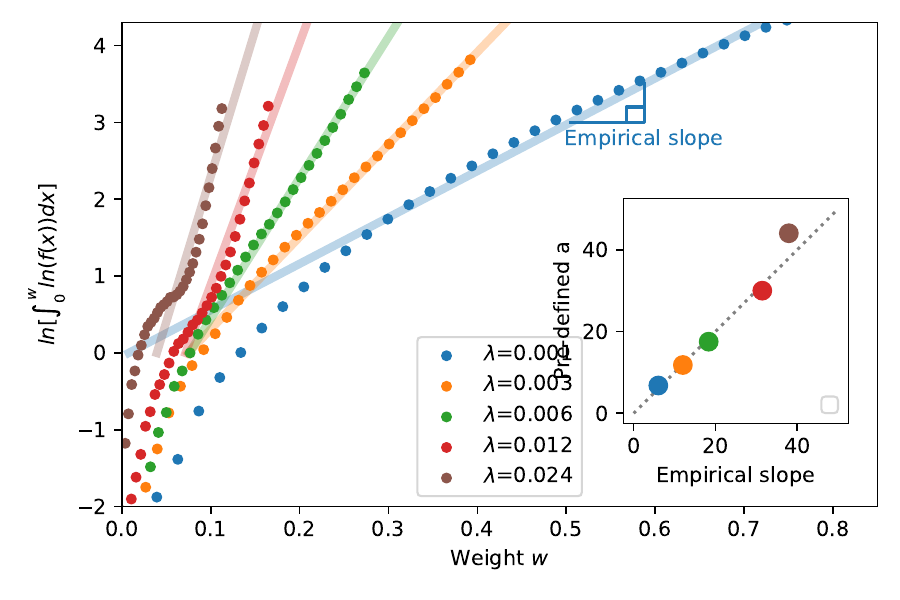}}
\caption{(a) $\ln\!\left[\int_0^w \ln(f(x)) dx\right]$ versus weight $w$ under different $\lambda$. (b) Comparison between empirically fitted slopes and predefined values of $a$.}
\label{b-vs-w}
\end{figure}

Figure~\ref{b-vs-w}(a) shows the transformed distribution along with the fitted linear model. Deviations at small $w$ arise from the approximation $e^{aw}-1 \approx e^{aw}$. To estimate the slope $a$ with less effect on these deviations, we apply a weighted linear fit, giving higher importance to larger-$w$ samples where the approximation is more accurate. The fitted line is shown as a shaded color in Figure~\ref{b-vs-w}(a).

To validate the estimation, Fig.~\ref{b-vs-w}(b) compares the empirically fitted slopes with the device-derived values. The results show strong agreement across most settings, with slight deviation at $\lambda=0.024$ due to imperfect fitting at the distribution tail. Overall, this confirms that the device-derived parameter $a$ is consistent with empirical observations.

\section{Accuracy vs. Number of Epochs}

\begin{figure}[!t]
\centering
\includegraphics[width=\linewidth]{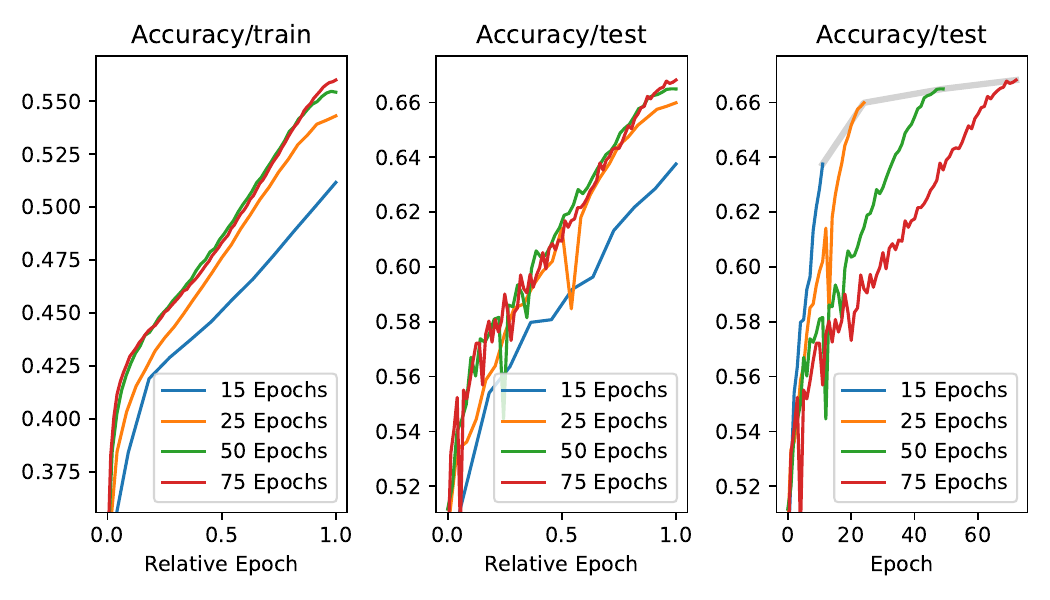}
\caption{Training and test accuracy versus number of epochs. Accuracy saturates after approximately 50 epochs.}
\label{fig:acc-vs-epoch}
\end{figure}

All ReRAM-aware finetuning experiments in this work are conducted for 50 epochs in image classification tasks. To justify this choice, Fig.~\ref{fig:acc-vs-epoch} shows the training and test accuracy as a function of training epochs.

Figures~\ref{fig:acc-vs-epoch}(a) and \ref{fig:acc-vs-epoch}(b) present the normalized training and test accuracy, respectively. The results indicate that most performance gains are achieved within the first 50 epochs, with only marginal improvements beyond this point.

Therefore, training beyond 50 epochs provides limited benefit while increasing computational cost, justifying the use of 50 epochs in our experiments.

\end{document}